\documentclass[11pt, a4paper, copyright, gdm]{google}

\usepackage[authoryear, sort&compress, round]{natbib}
\usepackage{wrapfig}

\usepackage{booktabs}  % For professional looking horizontal rules
\usepackage{tabularx}  % For auto-sizing columns
\usepackage{ragged2e}  % For better ragged text alignment in narrow columns
\usepackage{makecell}
\usepackage{subcaption}
\usepackage{graphicx}

\usepackage{xcolor}
\usepackage{todonotes} 

\uselogo{} 

\title{SIMA 2: A Generalist Embodied Agent for Virtual Worlds}

\author{SIMA Team, Google DeepMind\footnote{Please cite as SIMA Team, 2025 \\
Correspondence to: sima2-contact@google.com}}

\begin{abstract}
We introduce SIMA 2, a generalist embodied agent that understands and acts in a wide variety of 3D virtual worlds. Built upon a Gemini foundation model, SIMA 2 represents a significant step toward active, goal-directed interaction within an embodied environment. Unlike prior work (\textit{e.g.}, SIMA 1) limited to simple language commands, SIMA 2 acts as an interactive partner, capable of reasoning about high-level goals, conversing with the user, and handling complex instructions given through language and images. Across a diverse portfolio of games, SIMA 2 substantially closes the gap with human performance and demonstrates robust generalization to previously unseen environments, all while retaining the base model's core reasoning capabilities. Furthermore, we demonstrate a capacity for open-ended self-improvement: by leveraging Gemini to generate tasks and provide rewards, SIMA 2 can autonomously learn new skills from scratch in a new environment. This work validates a path toward creating versatile and continuously learning agents for both virtual and, eventually, physical worlds.
\end{abstract}

\begin{document}

\maketitle

\begin{figure}[h]
    \centering
    \includegraphics[width=\linewidth]{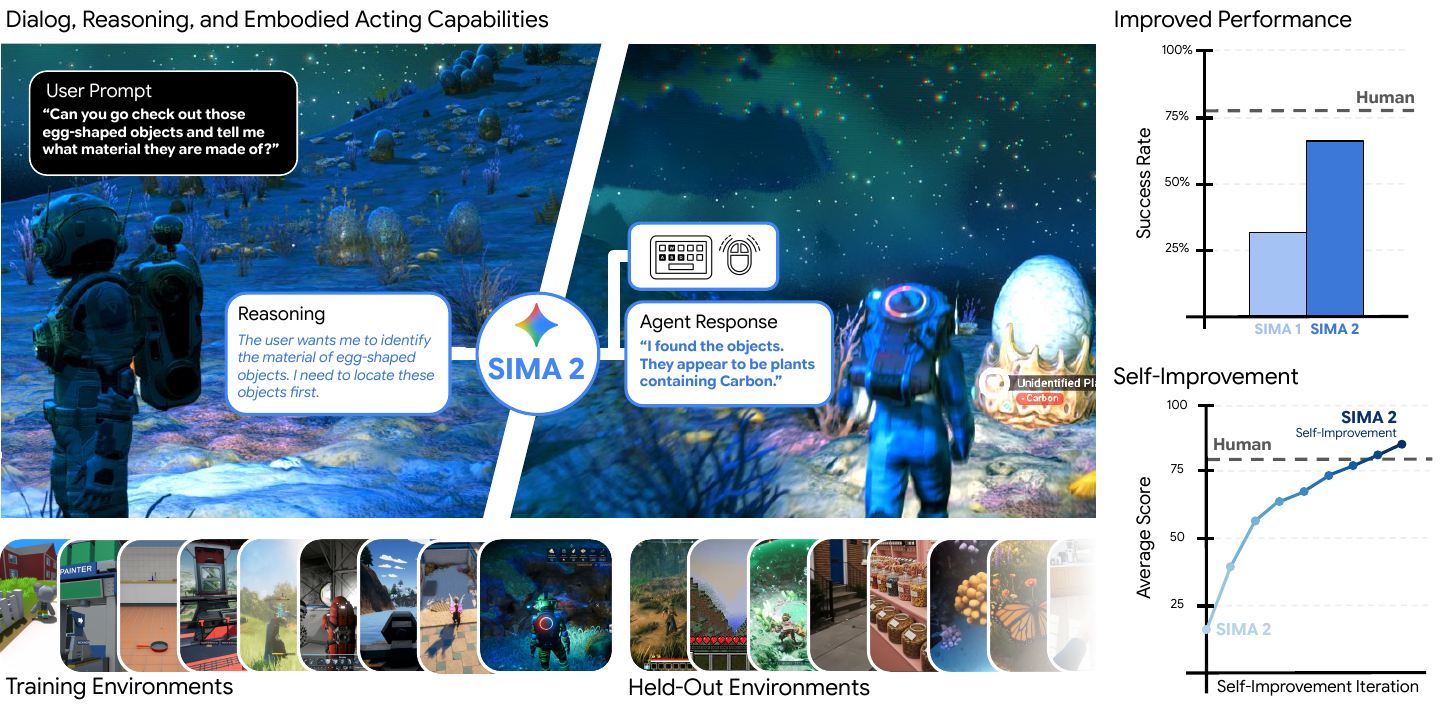}
    \caption{\textbf{SIMA 2} is a Gemini-based agent that reasons, acts, and engages in dialogue across diverse embodied 3D virtual worlds. In the top left panel, we see an example of the agent responding to the user in No Man's Sky. As compared with SIMA 1, SIMA 2 is a step-change improvement in embodied performance, and it is even capable of self-improving in previously unseen environments.}
    \label{fig:figure1}
\end{figure}

\section{Introduction}

Foundation models have achieved remarkable success in recent years \citep{anthropic2024claude3, openai2023gpt4, team2025gemini, bai2023qwen}, demonstrating a capacity for complex reasoning and understanding about the world. These models are primarily trained on vast amounts of static internet-scale datasets, allowing them to process and generate language, images, and video with impressive fluency. However, this results in an intelligence that is fundamentally disembodied and passive, leading to deficits in embodied performance noted in, \textit{e.g.}, \cite{majumdar2024openeqa}, \cite{yang2025embodiedbench}. They face a modern instantiation of Moravec’s Paradox: high-level cognitive tasks, such as playing chess or summarizing law, have proven easier to achieve than the low-level sensorimotor skills required to clear a dinner table or navigate a cluttered room \citep{moravec1988mind}.

The next great frontier for artificial intelligence is to move beyond passive understanding to active participation -- to create \textbf{foundation agents} that can operate within the embodied 3D worlds with a sense of agency, pursuing goals by learning to interact with their environment (\textit{c.f.} \cite{silver2025experience}), generalizing beyond limited scenarios, and displaying ``spatial intelligence'' \citep{gardner1983frames, feifei2025spatial}. In effect, this requires \textbf{embodiment}: the ability for an agent to go beyond merely perceiving the environment to also taking meaningful actions to change the state of that environment and learning from the resulting consequences. This is natively challenging for large language models (LLM) or vision-language models (VLM), as they were not trained to perform actions or understand the consequences of actions.

Our prior work, SIMA (Scalable Instructable Multiworld Agent) \citep{simateam2024scaling}, trained a single agent (henceforth referred to as SIMA 1) to follow hundreds of basic natural language instructions (\textit{e.g.}, ``Go to the campfire'') across a diverse set of 3D virtual games, demonstrating that it was possible to create a generalist agent that could operate and follow language instructions across many different worlds. These diverse and realistic simulations provide a scalable and safe testbed where an agent can learn fundamental embodiment capabilities by operating as a person does in these games: observing pixels on a screen and taking actions through a keyboard-and-mouse interface.  However, SIMA 1 was limited to short and direct instructions, could not respond in language or reason about its actions, and often displayed brittleness in generalizing to new situations or instructions.

Here we introduce SIMA 2, a step-change in embodied performance and capabilities. By integrating Gemini at its core, SIMA 2 moves beyond simple instruction-following to become a capable interactive companion. Where SIMA 1 needed to be told what to do step by step, SIMA 2 can reason about high-level goals, understand a user's intent, formulate multi-step plans, and converse about its strategy. This shifts SIMA 2 from reactive or low-level behavior to agentic, goal-oriented reasoning that is critical to more human-like forms of behavior and intelligence. By training across a growing portfolio of 3D games, the agent shows a remarkable capacity to generalize to previously unseen environments, including photorealistic worlds generated on-the-fly by Genie 3 \citep{genie3}. SIMA 2 also readily interfaces with more powerful Gemini models to enable even more advanced forms of reasoning and behavior. Finally, SIMA 2 is capable of open-ended self-improvement, learning new skills from its own experience, even in previously unseen environments. 

Collectively, these results validate the approach of incorporating embodied intelligence and agentic control within foundation models. By using diverse virtual worlds as a training ground, we see broad generalization and the capacity for further self-improvement. SIMA 2 thus represents a critical step toward creating general-purpose, interactive agents. It offers a promising path to eventually transferring these learned embodied capabilities to applications in the physical world, such as robotics.

\section{Background \& Related Works}

\paragraph{Games \& Simulation Driving Agent Research}

Our work builds on a long history of using games and simulation to drive agent research \citep{shannon1950programming, turing1953digital, samuel1959checkers}. In recent years, there has been an emphasis on increasingly complex games and simulations that more closely resemble the physical world. In the realm of simulation, physics engines like MuJoCo \citep{todorov2012mujoco} and other simulators \citep{coumans2016, beattie2016deepmind, kolve2017ai2, savva2019habitat, abramson2020imitating, makoviychuk2021isaac, deitke2022procthor} have been instrumental in driving progress in agents and robotics research. However, the complexity of these worlds is limited by the extent to which we can incorporate physical realism and other entities (objects, other agents, \textit{etc.}). Others have turned to video games as a source of complex worlds for agent research. Notably, \cite{bellemare2013arcade} established a suite of Atari games as environments for agent research, yielding breakthroughs in deep reinforcement learning \citep{mnih2015human, mnih2016asynchronous}. Similarly, OpenAI Universe \citep{openai2016universe} was intended as a platform of diverse, visually complex video games (though these were mostly 2D). Researchers eventually adopted more advanced games to train agents, moving to 3D \citep{kempka2016vizdoom, johnson2016malmo} and multi-agent games \citep{berner2019dota, vinyals2019grandmaster}. Of particular interest are \textit{open-world} games, like Minecraft, which require a broad range of skills \citep{guss2019minerl, baker2022video, fan2022minedojo, wang2023voyager, lifshitz2023steve, wang2023jarvis}. Much like the physical world, agents must learn to complete tasks in the absence of any clear, environment-provided reward, necessitating research on defining such goals and rewards \citep{fan2022minedojo, zhang2023omni}. More recently, with the move toward foundation models, several themes in games-driven agent research have emerged. There has been a push toward generalist agents \citep{reed2022generalist, lee2022multi, wang2025game, bytedance2025lumine, simateam2024scaling}, with a single agent tackling a range of skills across multiple game environments. Likewise, various works have explored the pursuit of long-horizon goals, such as completing entire MS-DOS and Game Boy games \citep{zhang2025videogamebench, hershey2025claude, zhang2025claude, team2025gemini}. Foundation models have also been benchmarked on reasoning in the game NetHack \citep{paglieri2025balrog} and in-context imitation learning in Atari \citep{ruoss2025lmact}. Finally, there has been a continued move toward more complex and visually-rich games, such as Counter-Strike \citep{pearce2022counter}, Red Dead Redemption \citep{tan2024towards}, and others \citep{simateam2024scaling, bytedance2025lumine, wang2025game, sharma2024toward}. SIMA 2 builds upon these themes, presenting a generalist agent capable of reasoning and acting in complex 3D environments. Indeed, with recent advances in world models (see below), we see that SIMA 2 is capable of generalizing \textit{beyond} video game environments to photorealistic worlds generated by Genie 3 \citep{genie3}.

\paragraph{World Models}

Along with training agents in virtual worlds, others have focused on \textit{learning} virtual worlds, sometimes referred to as world models. These models predict future outcomes based on current observations and actions. Early works described using such models for planning \citep{mel1987murphy, werbos1987learning, schmidhuber1990making}, exploration \citep{schmidhuber1991curious}, and offline learning \citep{sutton1990integrated}. However, it is only with recent advances in generative modeling that world models have proven capable of generating 3D visual observations \citep{ha2018world, valevski2025diffusion}. These models have similarly been demonstrated in the context of planning \citep{hafner2019learning}, exploration \citep{sekar2020planning, mendonca2021discovering}, and offline learning \citep{hafner2020dream}. More recently, these models have been applied to more complex environments, such as Minecraft \citep{hafner2025mastering} and Bleeding Edge \citep{kanervisto2025world, pearce2024scaling}. Beyond video games and simulation, world models have also been applied to real-world video for autonomous driving \citep{hu2023gaia, russell2025gaia}. While the aforementioned works modeled a finite set of environments, Genie \citep{bruce2024genie, parkerholder2024genie2} introduced a \textit{conditional} world model. By supplying a text description or initial observations, Genie is capable of generating limitless virtual worlds. We showcase SIMA 2 interacting with and self-improving in photorealistic environments generated by Genie 3 \citep{genie3}, demonstrating that SIMA 2 can generalize beyond video game environments. This points to a virtuous cycle between increasingly advanced world models and increasingly capable agents \citep{clune2019aigas}.

\paragraph{Foundation Models in Embodied Agents}

Early in the emergence of deep learning, embodied agents were largely trained from scratch \citep{mnih2015human, levine2016end, agrawal2016learning}. Accordingly, such agents largely failed to generalize outside of the settings in which they were trained \citep{kansky2017schema, huang2017adversarial}. To address this issue, researchers adopted pretrained visual representations, such as those derived from object classification \citep{pinto2016supersizing, zhu2017target, gupta2017cognitive} or contrastive pretraining \citep{shridhar2022cliport, nair2022r3m, simateam2024scaling}. Likewise, for language-conditional agents, researchers began adopting pretrained word embeddings \citep{anderson2018vision} and sentence embeddings \citep{lynch2020grounding, shridhar2022cliport} to enable broader generalization to new instructions. These approaches have ultimately culminated in embodied agents that are, \textit{themselves}, derived from pretrained foundation models \citep{driess2023palm}. Such ``vision-language-action'' (VLA) models \citep{brohan2023rt2} incorporate the benefits of large-scale internet pretraining into embodied agents, enabling generalization to novel objects and scenes. These agents have been applied to robotics \citep{black2024pi_0, intelligence2504pi0, team2025geminirobotics, team2025geminirobotics2, kim2024openvla} and virtual worlds \citep{bytedance2025lumine, hershey2025claude, zhang2025claude}, where integrating the reasoning capabilities of these models with embodied action has become an active area of research \citep{zhang2025embodied, sun2025emma, zhao2025cot}. Like these recent works, SIMA 2 is a VLA, containing a Gemini model \citep{team2023gemini, team2024gemini, team2025gemini} finetuned on data from 3D virtual worlds (\textit{c.f.}, \cite{team2025geminirobotics, team2025geminirobotics2}). Using virtual worlds as a testbed, we demonstrate SIMA 2's generalization capabilities, such as performing non-trivial tasks in new environments, including novel photorealistic environments. This broad generalization is in sharp contrast to the brittle initial generation of ``from-scratch'' agents \citep{kansky2017schema}, highlighting the field's progress toward achieving generalist embodied agents.

\paragraph{Open-Ended Self-Improvement}

A truly general embodied agent must possess the capacity to autonomously generate experience to drive adaptation and improvement. Indeed, a grand challenge of computer science is creating \emph{open-ended algorithms} \citep{stanley2015greatness, stanley2017open, clune2019aigas}, which produce never-ending innovation and learning. Current VLA agents, in contrast, are trained on datasets of existing demonstrations \citep{o2024open, kim2024openvla, intelligence2504pi0, team2025geminirobotics, bytedance2025lumine}. These works focus on a trained \textit{model} rather than a learning \textit{process}. Learning from experience has traditionally been the domain of reinforcement learning \citep{sutton1998reinforcement}. Yet, until recently, the field largely sidestepped two fundamental questions: 1) \textit{What outcome or goal (task) should be pursued?} and 2) \textit{How is progress toward this goal (reward) determined?} When confronting open-world environments, these questions become unavoidable. Various works have sought solutions to defining tasks and rewards. Early works used goal images \citep{zhu2017target, nair2018visual} and natural language goals \citep{mei2016listen, hermann2017grounded, luketina2019survey}. In defining reward functions for natural language goals, one set of approaches has used alignment between encoded language goals and visual inputs \citep{fan2022minedojo, ma2023liv, baumli2023vision, rocamonde2024vision, sontakke2023roboclip}. Other works have used foundation models to provide preference feedback \citep{wang2024rl, liu2025vlp}, programmatic rewards \citep{ma2023eureka, yu2023language, zhang2023omni}, or task completion estimates \citep{ghasemipour2025self, zhai2025vision}. With tasks and reward functions specified, the question then becomes \textit{which} tasks to pursue within an open-ended learning process. Many forms of goal-conditioned intrinsic motivation have been proposed \citep{colas2022autotelic}, yet one approach is to rely, again, on foundation models to provide novel, interesting tasks at the cusp of the agent's capabilities \citep{du2023guiding, zhang2023omni}. Like these previous works, we use foundation models both to propose tasks as well as to score the resulting trajectories. However, we do so in the context of training a VLA agent in novel 3D virtual worlds. By using three foundation models (task setter, agent, reward model), as well as a general world model, we demonstrate an open-ended self-improvement process capable of autonomously acquiring new skills in new environments.

\section{Methods}

\subsection{Environments}

\begin{figure}[t!]
    \centering
    \includegraphics[width=\linewidth]{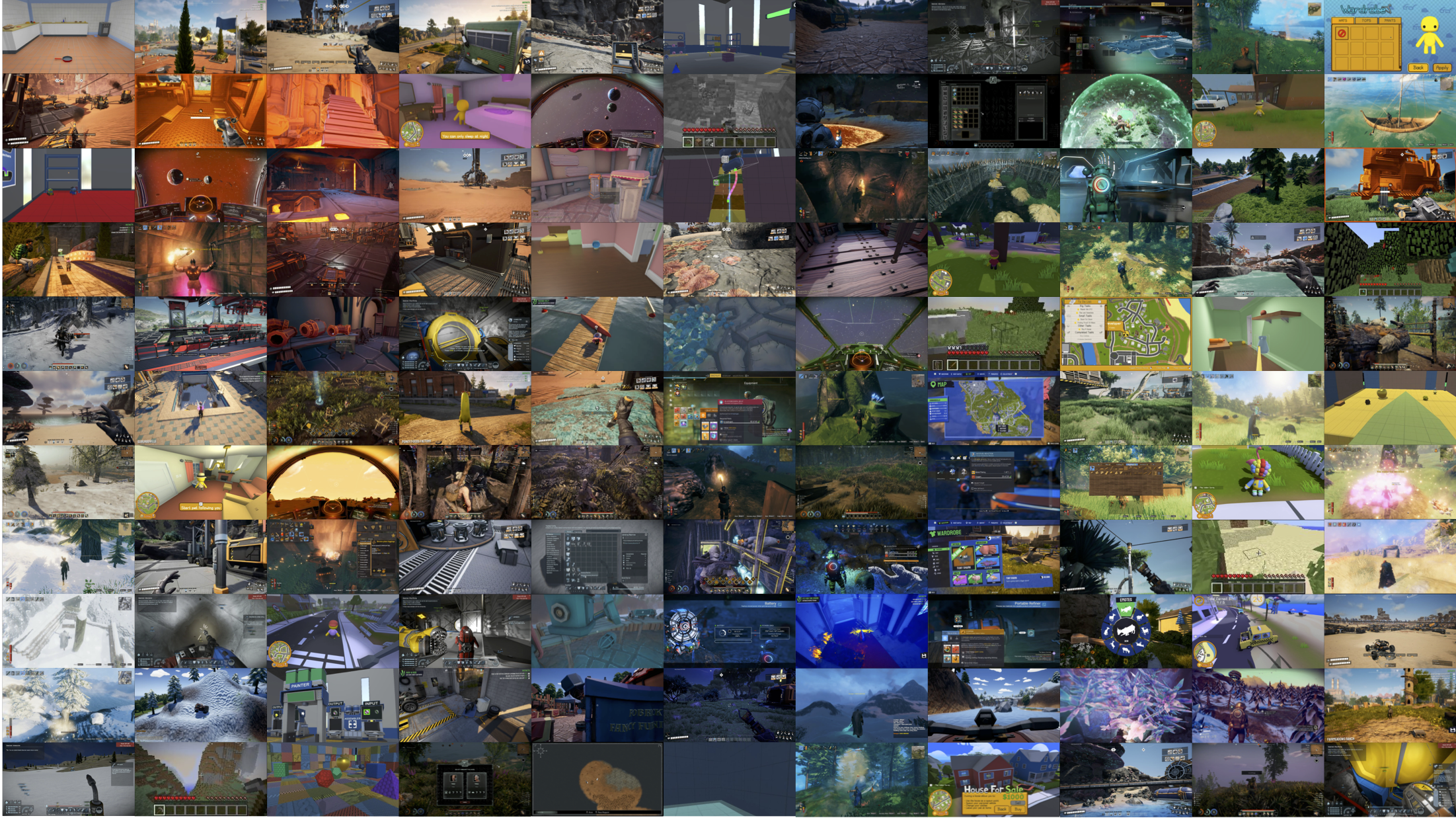}
    \caption{\textbf{Environments}. The grid shows a sampling of images across the video game environments used to train and evaluate SIMA 2. Due to the complexity of open-world commercial video games, agents must handle a near-limitless variety of 3D configurations, menus, and underlying environment dynamics. This provides an ideal setting to develop and test embodied agents. By acquiring general embodiment capabilities in these environments, SIMA 2 is able to generalize in non-trivial ways to entirely new environments, including photorealistic environments generated by Genie 3.}
    \label{fig:environments}
\end{figure}

As in SIMA 1, we use a combination of academic research environments and a variety of commercial video game environments licensed specifically for training and evaluating SIMA 2. For SIMA 2, we train agents on the research environments Construction Lab \citep{simateam2024scaling}, Playhouse \citep{abramson2020imitating}, and WorldLab \citep[\textit{e.g.},][]{paine2019making}, and the commercial video games Goat Simulator 3, Hydroneer, No Man's Sky, Satisfactory, Space Engineers, Valheim, and Wobbly Life (see \citet{simateam2024scaling} for an in-depth description of these environments). We further evaluate on a host of other games, including Minecraft, ASKA, and others (see Section \ref{sec:held_out_environments} for more details). A sampling of screenshots from these environments is shown in Figure \ref{fig:environments}. Of our training environments, Space Engineers is a newly-added environment since SIMA 1. We briefly describe this environment below.

\paragraph{Space Engineers} \href{https://www.spaceengineersgame.com/}{Space Engineers} is a sandbox game in which the player is an astronaut, using tools (drill, grinder, welder) to mine for resources and build voxel-based buildings and vehicles (ships, rovers, \textit{etc}.). Terrains include both asteroids and planets, with varying gravitational force. Additionally, the astronaut is equipped with a jetpack that enables motion along six degrees of freedom.

\subsubsection{Held-Out Environments}
\label{sec:held_out_environments}

Generalization is an important aspect of assessing agent capabilities, evaluating performance when confronted with novel situations. While the evaluations for all of our environments start from held-out \textit{states} (\textit{i.e.}, saved checkpoints) that are not present in the training data, many aspects of the environment are consistent, such as menus, maps, items, \textit{etc}. To assess a more extreme form of generalization, we also evaluate agents in entirely held-out \textit{environments}, where agents encounter new visuals, menus, and game mechanics. We quantitatively assess SIMA 2 on two held-out environments: ASKA and a subset of the MineDojo benchmark suite in Minecraft \citep{fan2022minedojo}. We also assess SIMA 2 qualitatively in The Gunk and a variety of Genie 3 \citep{genie3} environments.

\paragraph{ASKA}
\href{https://playaska.com/}{ASKA} is a Viking survival game in which the player builds a village, amassing villagers and assigning them to various tasks, \textit{e.g.}, harvesting wood or stone, farming, defenses, \textit{etc}. Despite differing visuals and mechanics, the game contains many of the high level skills found in our other environments, including resource gathering, menu use, tool use, crafting, building, and combat. ASKA provides a unique opportunity to assess generalization to unfamiliar environments. In particular, as ASKA is a recent game (\textit{Early Access} since June, 2024), it allows us to evaluate SIMA 2, and, by extension, Gemini, in an entirely new setting.

\paragraph{Minecraft (MineDojo)}
MineDojo \citep{fan2022minedojo} is a benchmark suite of language-conditional tasks in Minecraft built on the Malmo platform \citep{johnson2016malmo}. For SIMA 2, we use a subset of 50 programmatic tasks for a range of combat, mining, and crafting tasks drawn from the \textit{Combat}, \textit{Harvest}, and \textit{Tech Tree} task categories, each with 15 random seeds (\textit{i.e.},  environment configuration). Given the prevalence of Minecraft content, MineDojo offers an interesting test of embodied generalization, allowing us to evaluate the extent to which SIMA 2 can rely on Gemini's prior understanding of Minecraft visuals and terminology to complete novel embodied tasks.

\paragraph{The Gunk} 
\href{https://thunderfulgames.com/games/the-gunk/}{The Gunk} is an action-adventure platformer game in which the player is a scavenger that has just arrived on a new planet. The game follows a seven-chapter storyline around cleaning up black and red ``gunk'' from the planet using a handheld suction tool. Once the gunk in an area is cleared, the planet's wildlife is restored. This game is distinct from our other environments; it is story-driven rather than open-world and the visual appearance is quite dark. The main skills required for the initial portion of the game are navigation and tool use.

\paragraph{Genie 3} Genie 3 \citep{genie3} is a generative world model, enabling real-time interaction (via keyboard and mouse controls) with an endless number of newly-created environments. Environments can be conditioned using text descriptions or initial frames. For our evaluations, we generate a variety of photorealistic environments in a range of naturalistic and urban settings. These environments allow us to assess whether SIMA 2, by leveraging Gemini's world knowledge, is capable of generalizing beyond video game worlds to photorealistic environments. Further, because these are newly-generated environments, these scenes do not appear within the training datasets. The combination of SIMA and Genie gives a hint of the powerful possibility of creating open-ended algorithms that combine agents that learn forever in an infinite expanse of procedurally-generated environments \citep{clune2019aigas, wang2019paired, faldor2024omni}.

\begin{figure}[t]
    \centering
    \includegraphics[width=\linewidth]{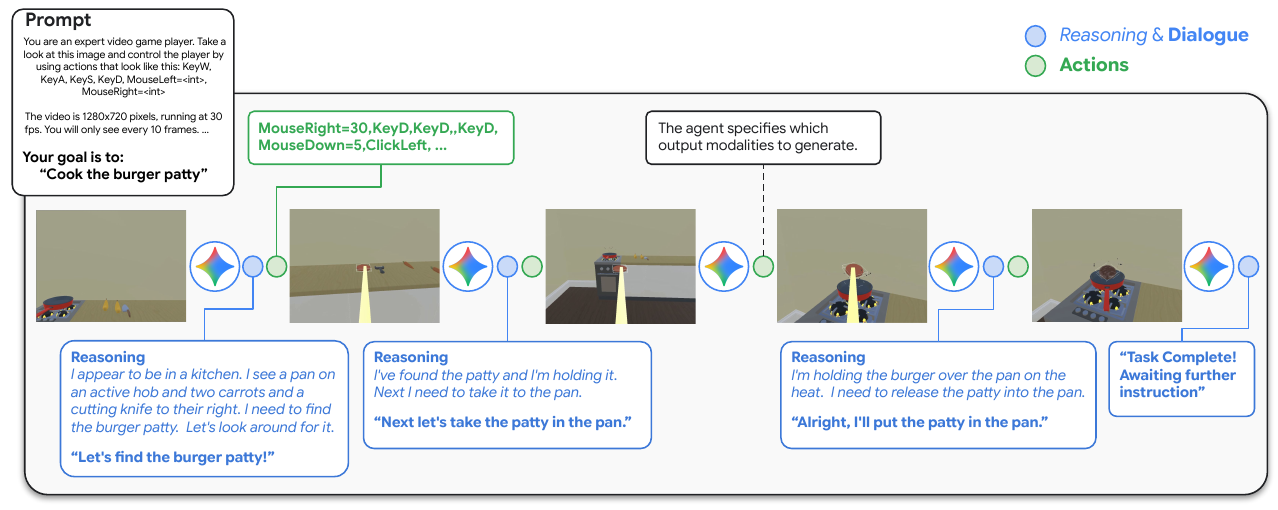}
    \caption{\textbf{Agent-Environment Interface}. The agent receives a prompt that includes the current instruction. Conditioning on recent frames, the agent outputs internal reasoning, dialogue, and actions, with the agent specifying which modalities to produce at any given step.}
    \label{fig:stream}
\end{figure}

\subsection{Agent-Environment Interface}

The agent-environment interface, shown in Figure \ref{fig:stream}, is designed to ensure that the agent perceives and acts within the game using the same modalities as a human player: visual input and keyboard-and-mouse actions. The agent does not receive any privileged information from the environment, such as an underlying state (\textit{c.f.}, \cite{hershey2025claude}). This interface manages the flow of information between the environment and the SIMA 2 agent.

The input to the agent consists of a stream of RGB video frames at a resolution of 720p. Periodically, the agent receives the latest frame from the environment and adds it to its history, which also includes the previous natural language inputs as well as the internal reasoning and responses produced by the agent (see Figure \ref{fig:stream}). The agent outputs chunks of actions that are then applied to the environment. The environmental action space emulates a standard human-computer interface, encompassing 96 standard keyboard keys, mouse clicks, and discretized mouse movements representing relative (x, y) position changes.
Instead of predicting discrete action tokens from a predefined set, the agent is trained via Supervised Fine-Tuning (SFT) to generate a structured text output. This output follows a specific format that can be deterministically parsed into low-level keyboard and mouse commands, as well as natural language for dialogue or internal reasoning.

\subsection{Data, Agent \& Training}
\label{sec:approach_training}

At its core, the SIMA 2 agent architecture is a Gemini Flash-Lite model that is trained using a mixture of gameplay and Gemini pretraining (non-gameplay) data.
We found this mixture crucial to maintain the original capabilities of the base model, such as vision understanding, dialogue, reasoning, and promptability. 
Starting from a pretrained Gemini Flash-Lite checkpoint, we perform supervised finetuning using this mixed dataset, training the model to produce keyboard-and-mouse action responses when prompted with image frames and an instruction.
The gameplay experience data includes two qualitatively different types of data:

\begin{itemize}
    \item \textbf{Human data} (Section \ref{sec:human_data}) are trajectories of post-processed human-collected data, which make up most of the training data by volume. They include text instructions together with the images captured from the environment and keyboard-and-mouse actions executed at each step. This type of data is crucial to teach the agent low-level acting and motor control in 3D environments.
    \item \textbf{Bridge data} (Section \ref{sec:bridge_data}) contain extra high-level interaction data between the user and the agent, such as dialogue and reasoning. This is synthetically generated using a Gemini model. Bridge data teaches the agent how to relate high-level instructions and dialogue from the user with internal reasoning and low-level actions. 
\end{itemize}

\subsubsection{Human Data}
\label{sec:human_data}

To train an agent that can simultaneously act, follow natural language instructions, and reason, we constructed a large-scale, multi-modal dataset that captures the richness of human gameplay in 3D environments. The main training dataset is composed of RGB video frames of gameplay, corresponding keyboard-and-mouse actions, and a variety of language annotations. This was generated primarily by human participants interacting with the games under licensed agreements. This is supplemented with synthetically-annotated data from Gemini to further scale our efforts. All participants provided informed consent prior to completing tasks and were reimbursed for their time. Datasets were collected using several methods to capture a wide range of behaviors and for various uses.

\paragraph{Gameplay Demonstration Data}
The bulk of our training data consists of gameplay demonstrations collected through two different approaches:
\begin{itemize}
\item \textbf{Single-person, post-hoc annotation:} In this approach, a single participant plays in a free-form manner, typically starting from the game's standard starting point. The recorded gameplay is later annotated in natural language by the player describing their actions, aligned to specific frames. While this method allows for the collection of diverse and naturalistic behavior, the language annotations are not causally tied to the player's intent, as they were constructed in hindsight.
\item \textbf{Two-person gameplay annotation (``Setter-Solver''):} To create a tighter causal link between language and action, we used a two-player interactive setup. One person, the ``Setter,'' watches the gameplay and issues live instructions to the other participant, the ``Solver,'' who controls the game avatar. Under this approach, the language instruction always precedes the corresponding actions, resulting in a more causally correct form of annotation than the single-person approach. Note that the Setter was only able to control the game avatar indirectly via the Solver following their instructions.
\end{itemize}

\paragraph{Task-Specific \& Evaluative Data}
In addition to open-ended gameplay, we collected data for predefined tasks and evaluations.
\begin{itemize}
\item \textbf{Episodic, task-specific scenarios (``Game-Tasks''):} To gather data and examples of specific skills, we created a framework for ``game-tasks,'' in which players are presented with a specific instruction (\textit{e.g.}, ``Craft a stone axe'') starting from a predefined game state. These episodes ended at either a prespecified time limit or when the player determined they had succeeded at the task, thereby ending the episode.
\item \textbf{Human ratings and comparisons:} To evaluate agent performance and calibrate reward models, we collected human judgments of previously collected game trajectories (typically collected in the ``game-task'' framework) to determine whether the player succeeded in the given task instruction. This includes binary success ratings for game-tasks as well as side-by-side comparisons of two separate trajectories to determine which more successfully accomplished a given task instruction.
\end{itemize}

\paragraph{Quality Assessment, Pre-processing, and Filtering}
Before data collection, human participants were given guided tutorials detailing the general game controls and mechanics, how to operate the game collection user-interface, as well as how to annotate the data with language labels or provide language instructions.
Prior to model training, we carry out several offline pre-processing steps. These include reshaping or resizing image frames to match what is expected for model input, employing various heuristics and score metrics to filter out low-quality data, and remixing and weighting data from different environments and datasets to optimize skill learning.
For the bulk of the data, we converted gameplay trajectories into ``spans,'' which entailed splitting them into shorter sub-sequences, each with a single task instruction. A span thus consists of a single task instruction that is associated with a sequence of video frames and actions taken during those steps.
Synthetic labeling was also applied offline by Gemini models to provide augmented language and reasoning text.

\subsubsection{Bridge Data}
\label{sec:bridge_data}

Human gameplay does not directly contain reasoning and dialogue. Thus, in order to train agents that can simultaneously act, reason, and engage in dialogue, we require some form of augmented data that combines these modalities. In particular, we require training data that interleaves reasoning and dialogue content that is consistent with the visual input and actions \citep{zhang2025embodied, sun2025emma, zhao2025cot}, similar to the format shown in Figure \ref{fig:stream}. Training our agent to respond in this way enables us to combine Gemini's vision and language understanding capabilities with embodied interaction.

To create this dataset, we select a relatively small number of high-quality data examples, featuring a variety of in-game behavior across all of our training environments. Each example contains a single task instruction and a sequence of actions and visual frames consistent with the successful completion of the task. Using Gemini Pro, we annotate each example with internal reasoning and dialogue in a manner that is causally consistent with the observable scene from the agent's ego-centric perspective and embodied behavior. We also vary the training prompt within these examples to induce additional robustness. The resulting examples contain a range of capabilities, including error correcting behavior, explicit instruction following, instruction chaining (\textit{i.e.}, following a sequence of instructions), visual question answering, reliance on memory, and long-horizon behavior. We also include \textit{no-ops} (time steps at which no actions are taken) to ensure that the agent remains still after a task has been completed. We refer to the resulting dataset as \textit{``bridge''} data, as these examples bridge the modalities of embodied action and language.

\subsubsection{Reinforcement Learning}
\label{sec:reinforcement_learning}

After the initial supervised learning stages, the agent is further trained using online reinforcement learning from verifiable rewards (\textit{c.f.}, \cite{wen2025rlvr, mankowitz2023alphadev}). To do this, we curated a set of verifiable tasks, \textit{i.e.}, a tuple of an initial game state, a text instruction, and a verification function. We then generate agent trajectories on these tasks in order to improve the policy. Reward is obtained for either successfully completing the embodied task or giving a correct answer to a question grounded in the environment. Some tasks contain additional shaped rewards to improve the instruction-following capabilities and controllability of the agent. 

The main body of tasks were collected from participants contracting with Google. Participants were placed into random game states and asked to explore the nearby environment and suggest multiple tasks that could be completed from that point. This set of tasks was expanded by applying verifier functions to all human trajectories (see Human Data section) to identify goal completion points and pairing these points with a nearby game state. These tasks were filtered down to those that a human could complete within a specified time limit to remove excessively hard tasks. In addition, we also generated dialogue tasks by selecting random screenshots from our human data and pairing these with human-suggested question-answer pairs.

This phase of RL training is limited to our training environments and excludes our held-out environments, such as ASKA and MineDojo.

\subsection{Evaluations}

Our quantitative analysis focuses on embodied tasks, in which the agent is given a text-based instruction and executes a series of keyboard-and-mouse actions in the environment to achieve a goal. As in \cite{simateam2024scaling}, task success is measured using one of three distinct types of evaluation function. We refer to the first two categories collectively as \textit{automatic} evaluations, as they do not require manual assessment:
\begin{itemize}
    \item \textbf{Ground-Truth Evaluation}: These evaluations use ground-truth state information from the environment to assess task success. For instance, success may depend on the absolute or relative positions of objects (``\textit{Lift the cube}''), the acquisition of an object or resource (``\textit{Gather wood}''), or triggering some other game mechanic (``\textit{Water the plant}''). Given that commercial video games do not generally expose this state information in an accessible way, these evaluations are limited to our research environments.
    \begin{itemize}
        \item \small \textit{Construction Lab, MineDojo, Playhouse, WorldLab}
    \end{itemize}
    \item \textbf{Programmatic Evaluation}: For commercial video games, which do not generally expose ground-truth state information, we define programmatic evaluations based on the game screen and the agent's keyboard-and-mouse actions. Video games often contain on-screen text in the form of pop-ups and menus, signaling events and state information. As in previous works \citep{openai2016universe, simateam2024scaling}, we use optical character recognition (OCR) to detect this on-screen text to determine task success. We also define functions over pixel colors and action outputs. While creating these task functions is a manual process, once written, they can be deployed easily at-scale to enable automatic evaluation across our commercial video game environments. However, these tasks are restricted to the outcomes that can be detected through heuristics over the visual input or through the agent's actions.
    \begin{itemize}
        \item \small \textit{ASKA, Goat Simulator 3, Hydroneer, No Man's Sky, Satisfactory, Space Engineers, Valheim, Wobbly Life}
    \end{itemize}
    \item \textbf{Human Evaluation}: For tasks where no ground-truth or programmatic task function can easily be written, we rely on human raters to assess task completion by observing the video of the agent's trajectory. As raters do not always agree on task success, we obtain five independent ratings per video to improve precision. Although this method is more costly than the other two (automatic) evaluations, it can be applied to a broader variety of tasks.
    \begin{itemize}
        \item \small \textit{Goat Simulator 3, Hydroneer, No Man's Sky, Satisfactory, Valheim, Wobbly Life}
    \end{itemize}
\end{itemize}
\paragraph{SIMA Evaluation Suite 2.0}
Since announcing SIMA 1, we have significantly expanded our evaluations. This includes expanding our programmatic and human evaluations to additional domains, increasing the number of evaluation tasks, often by an order of magnitude or more in the case of programmatic evaluations, and improving our programmatic evaluations to better align them with our expectations of task success. Here, we highlight three improvements.
\begin{itemize}
    \item Rather than triggering success after the first text detection, for applicable tasks, we ensure that the text is present for several seconds. By requiring a degree of \textit{persistence}, our evaluations select for more intentional behavior, demonstrated by the agent pausing when it considers a task to be completed.
    \item For a stricter constraint, in a subset of tasks we place a threshold on the number of actions permitted to be performed after task completion. Measuring whether agents remain still allows us to gauge whether agents recognize task completion and whether tasks can be readily chained during deployment.
    \item We have greatly expanded our set of sequential programmatically-evaluated tasks, in which each instruction is supplied once the previous task has been completed, closer reflecting the behavior expected in an interactive session. Succeeding on these tasks requires successfully completing every sub-task in the sequential chain. 
\end{itemize}
As a result of these improvements, our evaluations are substantially more challenging than those originally reported in \cite{simateam2024scaling}. Accordingly, the SIMA 1 agent obtains lower success rates.

\paragraph{Human Baselines}

To contextualize SIMA 2's performance, we established human baselines by collecting gameplay trajectories on our full evaluation suite of tasks. These were designed to closely replicate the agent's testing conditions, including the time limits for each task. For tasks in which the agent receives multiple instructions in a sequence, the players were given all steps to accomplish at once, with the guidance that they were to complete them one at a time in order. 

To ensure a representative and reliable human baseline, for our training environments we collected this data from players who had prior experience with the game through their participation in our training data collection. For the held-out environments, ASKA and MineDojo, we recruited new participants with general video game experience but no prior experience playing these specific titles. They were provided with written instructions on core game mechanics and controls but received no task-specific guidance.

\section{Results}

\subsection{New Capabilities}
\label{sec: results_new_capabilities}

Despite SIMA 1's ability to perform a broad range of short-horizon embodied tasks, it was also limited in several aspects. While SIMA 1 used pretrained vision encoders \citep{bica2024improving, villegas2022phenaki}, its language encoding was trained from scratch. Thus, SIMA 1's instruction-following capabilities were constrained to the vocabulary of the annotated gameplay on which it was trained. Further, SIMA 1 only mapped text instructions and current images to keyboard and mouse actions; it was incapable of processing any other inputs or outputs. For instance, SIMA 1 was incapable of outputting text (\textit{e.g.,} internal reasoning or dialogue), and it was also incapable of receiving multi-modal instruction prompts (\textit{e.g.,} sketches).

SIMA 2 overcomes these limitations, enabling a new set of capabilities. By powering SIMA 2 with Gemini, we inherit Gemini's broad vision and language understanding capabilities, opening up a far broader and more complex set of tasks. As vision, language, and action now occupy a single stream of tokens, we can lift the input and output constraints on the agent to enable new forms of interaction, such as dialogue and multi-modal prompting. This also enables internal reasoning, allowing the agent to adapt its behavior based on internal inferences. Below, we highlight several of these new capabilities, paired with concrete examples.

\begin{figure}
    \centering
    \includegraphics[width=\linewidth]{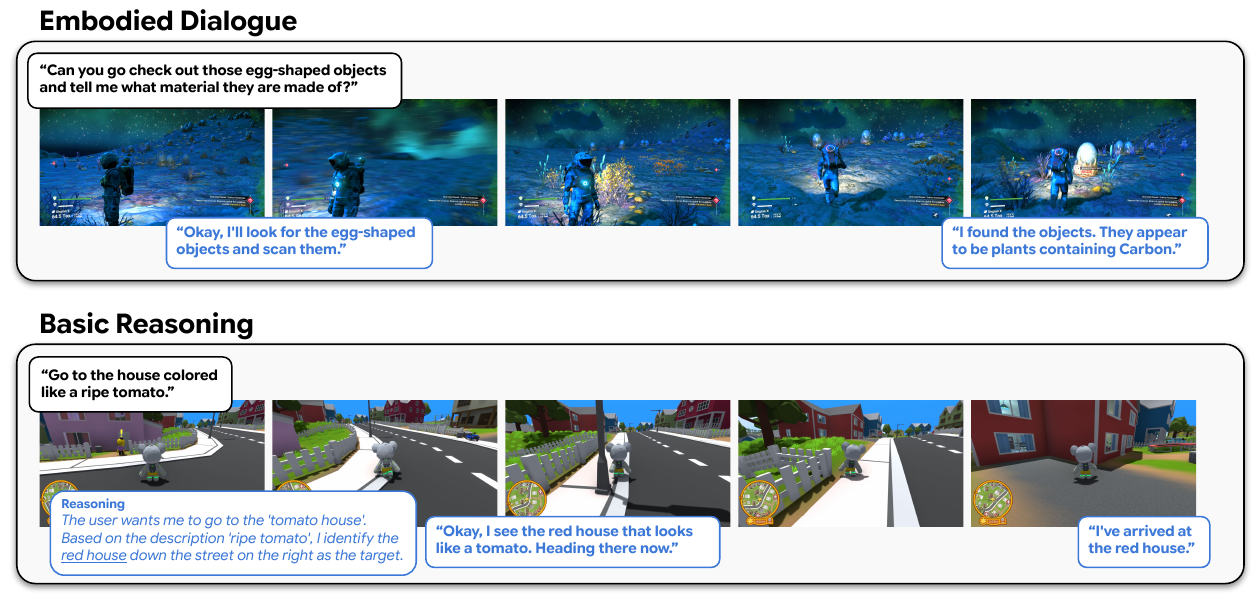}
    \caption{\textbf{Embodied Dialogue \& Basic Reasoning}. SIMA 2 contains a variety of new capabilities, including embodied dialogue and basic reasoning. Above, SIMA 2 answers a user's question through embodied interaction. Below, the agent correctly reasons that it needs to go to a \textit{red} house based on the user's instruction. These new capabilities are unlocked by using Gemini within SIMA 2.}
    \label{fig:new_capabilities1}
\end{figure}

\paragraph{Embodied Dialogue}
SIMA 2 is, at its core, a Gemini model. Thus, just like Gemini, it can engage in dialogue with a user, making use of Gemini's general world knowledge and visual question-answering capabilities. However, because SIMA 2 is situated in a 3D world, it can also take actions in response to user prompts, enabling a new capability for \textit{embodied dialogue}. This covers a wide variety of interactions, including confirmations of users' requests and proactively responding when tasks have been completed. SIMA 2 can even ask clarifying questions when a user's request is ambiguous. One particularly unique form of interaction is embodied question-answering, in which a user asks or instructs the agent to find some piece of information, to which SIMA 2 must take embodied actions to determine the answer and respond in natural language. For instance, in No Man's Sky, when asked \textit{``Can you go check out those egg-shaped objects and tell me what material they are made of?''}, SIMA 2 confirms the user's instruction, then navigates to one of the objects, using the on-screen text to reply \textit{``I found the objects. They appear to be plants containing Carbon.''} This example highlights SIMA 2's ability to engage in embodied information-seeking behavior, going beyond the capabilities of SIMA 1 and the base Gemini model.

\paragraph{Basic Reasoning}

In the same way that SIMA 2 can output text externally to engage in dialogue with a user, it can also output text \textit{internally} to perform reasoning. By generating and conditioning on internal reasoning, SIMA 2 can use internal inferences to modify its own behavior. With Gemini's general reasoning capabilities, SIMA 2 can thus handle more indirect, nuanced, or novel instructions, which are not present in the training data. To provide a simple, illustrative example, a user can instruct the agent to \textit{``Go to the house colored like a ripe tomato.''} Internally, the agent then reasons, \textit{Based on the description ``ripe tomato'', I identify the red house down the street on the right as the target.} The agent then responds and heads to the correct house. This general ability to modify behavior based on internal reasoning affords a broad array of novel behaviors. Indeed, as will be shown in Section \ref{sec: out-of-domain performance}, SIMA 2 uses its internal reasoning to correctly identify appropriate actions in entirely novel environments. 

\begin{figure}
    \centering
    \includegraphics[width=\linewidth]{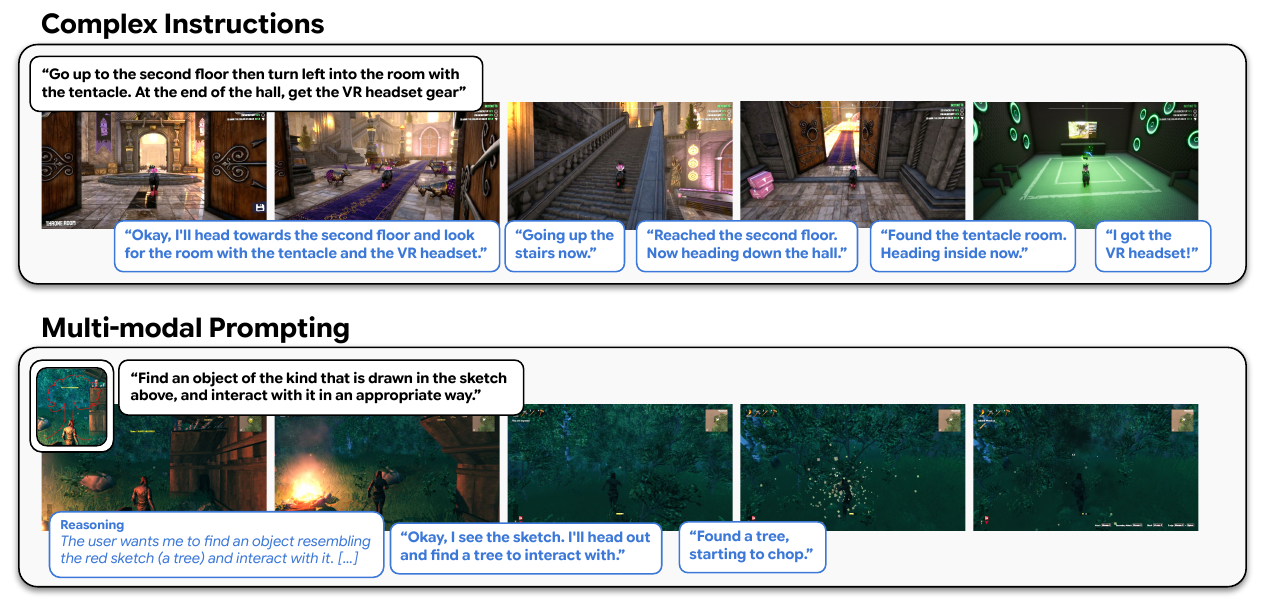}
    \caption{\textbf{Complex Instructions \& Multi-modal Prompting}. By inheriting Gemini's language understanding capabilities, SIMA 2 can handle a variety of novel, complex instructions, including breaking down instructions to successfully navigate to a specific room. SIMA 2 can also be prompted with images, including sketches, to specify locations, paths, or objects.}
    \label{fig:new_capabilities2}
\end{figure}

\paragraph{Complex Instructions}

SIMA 2 also benefits from Gemini's general language understanding capabilities, allowing it to generalize to novel, complex instructions. For instance, by leveraging the zero-shot multilingual capabilities inherited from the base model, SIMA 2 can readily perform tasks when instructed in French, German, Mandarin Chinese, \textit{etc}, despite only training on embodied data in English. SIMA 2 can even interpret instructions provided in emojis, \textit{e.g.}, inferring that an ax emoji and a tree emoji implies chopping down a tree. This also extends to more complex, multi-step instructions. For instance, when given the navigation instructions \textit{``Go up to the second floor then turn left into the room with the tentacle. At the end of the hall, get the VR headset gear''}, the agent can successfully go through each step, reporting its progress along the way. Without dedicated training data, a task of this form would be far outside of the scope of SIMA 1. In contrast, this ability to more fully utilize language to perform embodied tasks means that SIMA 2 can better harness the abstract, compositional properties of language.

\paragraph{Multi-modal Prompting}

Gemini is natively multi-modal, processing images, audio, and video in addition to text. SIMA 2 thus inherits multi-modal prompting capabilities, allowing us to instruct the agent in novel ways. In our investigations, we have primarily focused on images, as they offer a simple way to transcend the limitations of language instructions. Such images, for instance, can come from game wikis or even generative image models. One particularly helpful use-case is \textit{sketching}; rather than describing a location, a path, or an object in text, we can simply annotate the current game image to indicate it. We can even draw an object on the game screen and provide this in the instruction. For instance, when given a sketch of a tree and instructed, \textit{``Find an object of the kind that is drawn in the sketch above, and interact with it in an appropriate way.''}, the agent correctly identifies it as a tree, then proceeds to chop one down. Thus, these new ways of instructing the agent enable new forms of more intuitive interaction.

\subsection{Embodied Task Performance}
\label{sec: results_embodied_task_performance}

\begin{figure}
    \centering
    \includegraphics[width=0.75\linewidth]{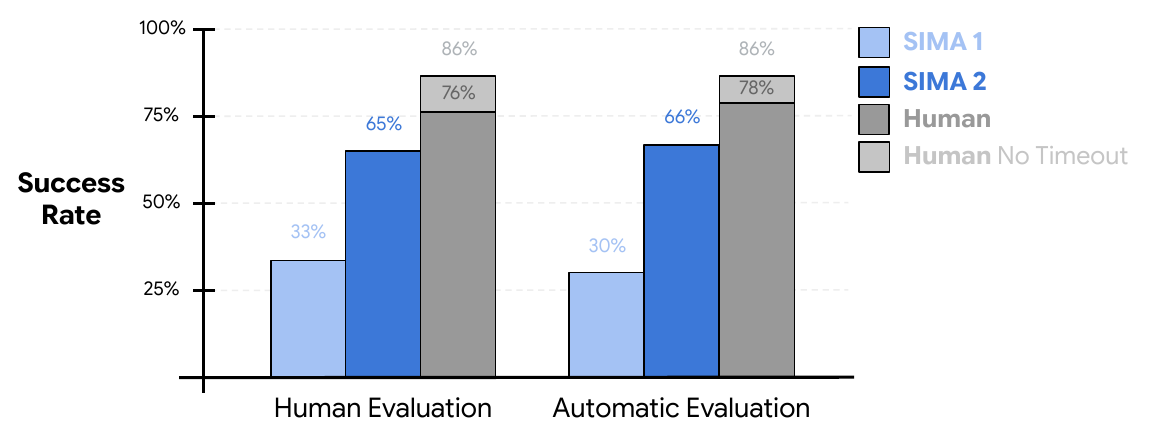}
    \caption{\textbf{Average Performance on Embodied Tasks}. Performance is averaged over training environments for each type of evaluation (human or automatic). We plot human performance both with and without the time restrictions imposed on agents. SIMA 2 effectively doubles the average success rate of SIMA 1, approaching human-level performance in both cases.}
    \label{fig:average_performance}
\end{figure}

\subsubsection{Performance in Training Environments}

\begin{wrapfigure}{l}{0.6\textwidth} % 'l' for left, '0.5\textwidth' is the width reserved for the image
    \centering
    \includegraphics[width=\linewidth]{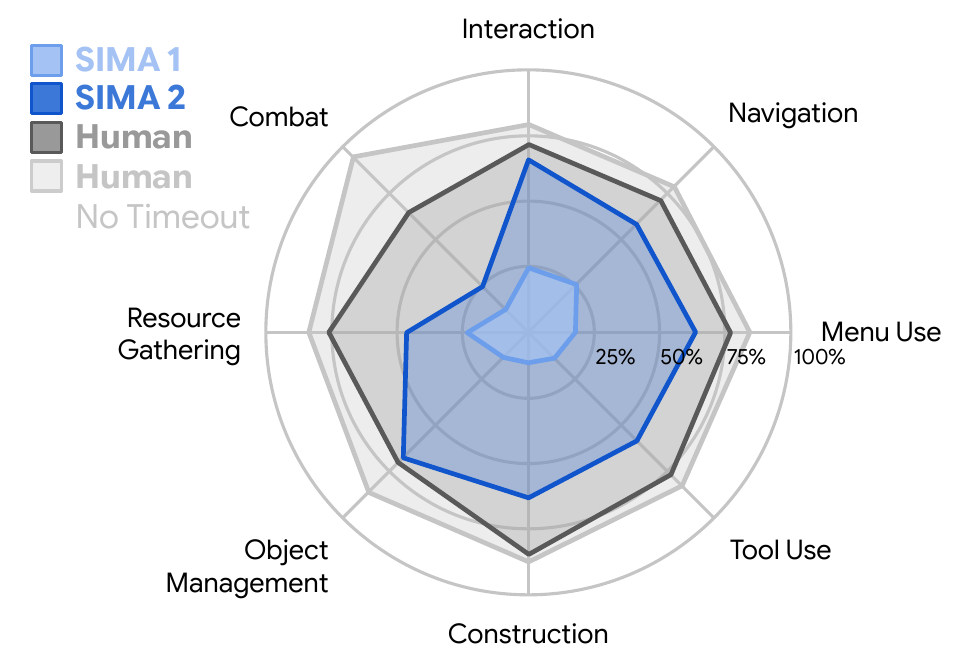}
    \caption{\textbf{Performance Across Skill Categories}. SIMA 2 significantly improves over SIMA 1 across multiple skill categories. In categories like interaction and object management, SIMA 2 nearly closes the gap with human-level performance. However, in other categories, like resource gathering and combat, SIMA 2 still has room for improvement.}
    \label{fig:performance_by_skill}
\end{wrapfigure}

In Figure \ref{fig:average_performance}, we plot the performance of SIMA 1, SIMA 2, and humans across our human-evaluated and automatically-evaluated embodied task sets, averaged over environments seen during training. Human baseline performance is collected from individuals with significant gameplay experience in each environment. We plot human performance subject to the same task timeouts given to agents (dark gray), as well as performance without this restriction (light gray). The latter value gives an approximate upper bound on performance, as we observed that participants frequently struggled to complete tasks within the allotted time, some of which were as short as three seconds. Primary sources of difficulty for human players included initial inattention, infrastructure latency, and challenges with the fine-motor control required to operate game interfaces as fluidly as the agent.

Overall, we find that SIMA 2 substantially outperforms SIMA 1, effectively doubling the success rate and nearly closing the gap with human performance. This is remarkably consistent across both human-evaluated and automatically-evaluated tasks.

\begin{figure}
    \centering
    \includegraphics[width=\linewidth]{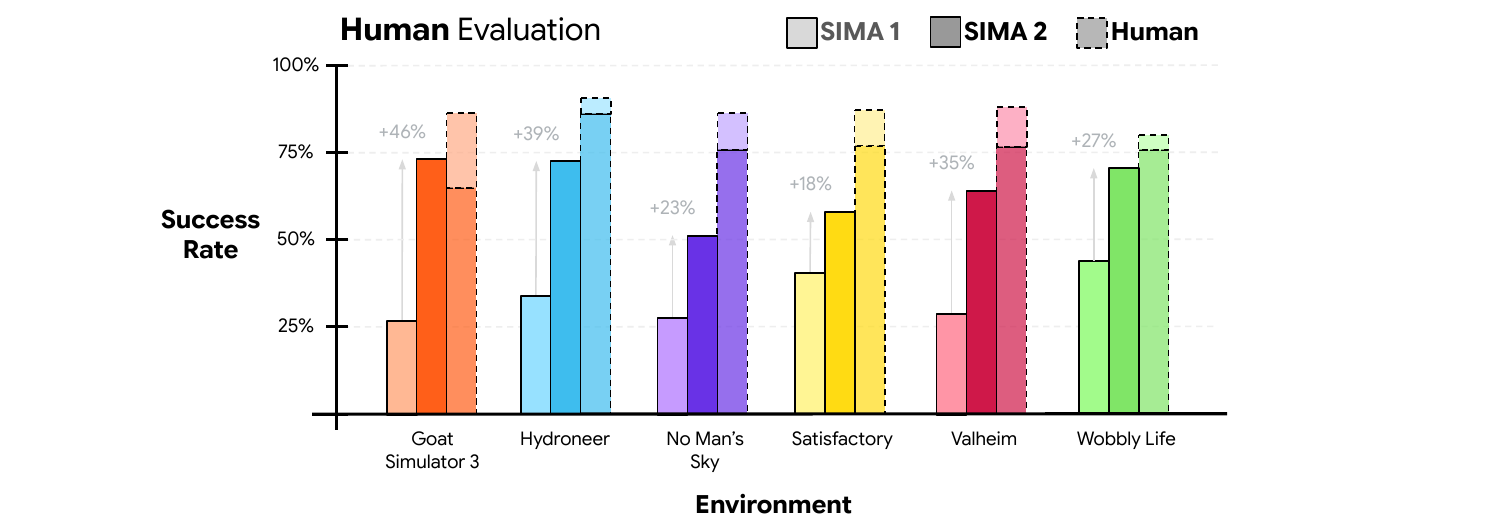}
    \caption{\textbf{Performance on \textit{Human} Evaluations By Environment}. Bars show the task success rate of SIMA 1, SIMA 2, and humans on a suite of tasks from training environments, with success measured by 5 independent human ratings per trial. Human performance is plotted subject to the same timeout restrictions as imposed on agents (darker) and with this restriction removed (lighter). SIMA 2 improves significantly over SIMA 1 across all environments, nearly closing the gap with human performance in many cases.}
    \label{fig:human_evaluations_by_domain}
\end{figure}

\begin{figure}
    \centering
    \includegraphics[width=\linewidth]{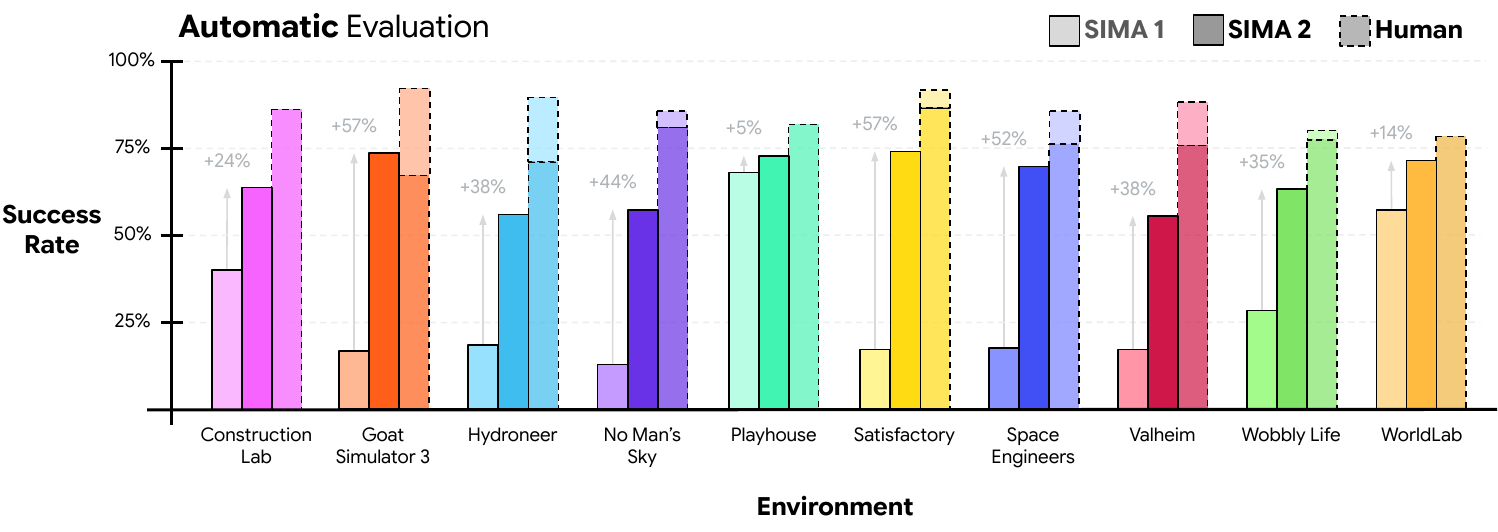}
    \caption{\textbf{Performance on \textit{Automatic} Evaluations By Environment}. Bars show the task success rate of SIMA 1, SIMA 2, and humans on a suite of tasks from training environments, with success measured by ground truth rewards (Construction Lab, Playhouse, and WorldLab) or programmatic evaluation (all other environments). Where applicable, human performance is plotted with and without the same timeout restrictions as imposed on agents. SIMA 2 improves significantly over SIMA 1 across nearly all environments, almost closing the gap with human performance in many cases.}
    \label{fig:automatic_evaluations_by_domain}
\end{figure}

In Figures \ref{fig:human_evaluations_by_domain} \& \ref{fig:automatic_evaluations_by_domain}, we break down agent and human performance across each environment. Where applicable, we plot human performance both with and without timeout restrictions. SIMA 2 improves over SIMA 1 across all environments, nearly closing the gap with human performance in many cases. In particular, there are substantial performance improvements in video game environments. This highlights how SIMA 2 is better able to handle more complex settings, where agents are required to deal with visual diversity, interact with menus, and navigate more challenging game dynamics.

To better understand the performance improvements of SIMA 2, we decompose our evaluation tasks into skill categories, as previously performed in \citet{simateam2024scaling}. In Figure \ref{fig:performance_by_skill}, we plot the average performance across a subset of eight common skill categories: interaction, navigation, menu use, tool use, construction, object management, resource gathering, and combat. In Table \ref{tab:skill_categories} in Appendix \ref{appendix: embodied skill categories}, we briefly describe each category, along with several representative examples.

From Figure \ref{fig:performance_by_skill}, we see that SIMA 2 substantially improves over SIMA 1 across skill categories, approaching human-level performance in several instances. Notably, SIMA 2 still struggles with \textit{Combat}, in part, due to the motor difficulty of these tasks. For instance, hunting a deer in Valheim typically requires approaching from downwind while crouching, then quickly attacking. If the deer escapes, a challenging chase then ensues, requiring split-second decision making and a degree of luck. Similarly, when removing the timeout restriction (imposed on the agents), human performance in this skill category improves substantially.

\subsubsection{Performance in Held-Out Environments}
\label{sec: out-of-domain performance}

\begin{wrapfigure}{r}{0.48\textwidth} % 'l' for left, '0.5\textwidth' is the width reserved for the image
    \centering
    \includegraphics[width=\linewidth]{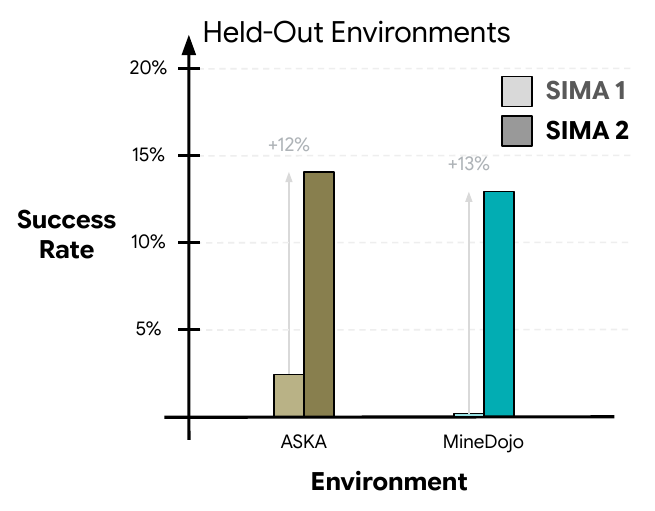}
    \caption{\textbf{Held-Out Environment Performance}. SIMA 2 outperforms SIMA 1 on two held-out environments (\textit{i.e.}, unseen during training): ASKA and MineDojo. This demonstrates that SIMA 2 is a more general agent, capable of performing non-trivial tasks in new settings.}
    \label{fig:ood_performance}
\end{wrapfigure}

The previous section discussed how SIMA 2 outperforms SIMA 1 on held-out \emph{tasks} within environments seen during training. This provides compelling evidence that SIMA 2 is a more \textit{performant} agent. We now address whether SIMA 2 is also a more \textit{general} agent. That is, can it generalize to new visual settings, menus, and game dynamics? To assess a more extreme form of generalization, we evaluate SIMA 2 on entirely held-out environments, previously unseen during training. We first present a quantitative evaluation comparing SIMA 1 and SIMA 2, then provide qualitative examples of SIMA 2's behavior in several wildly different environments.

\begin{figure}
    \centering
    \includegraphics[width=\linewidth]{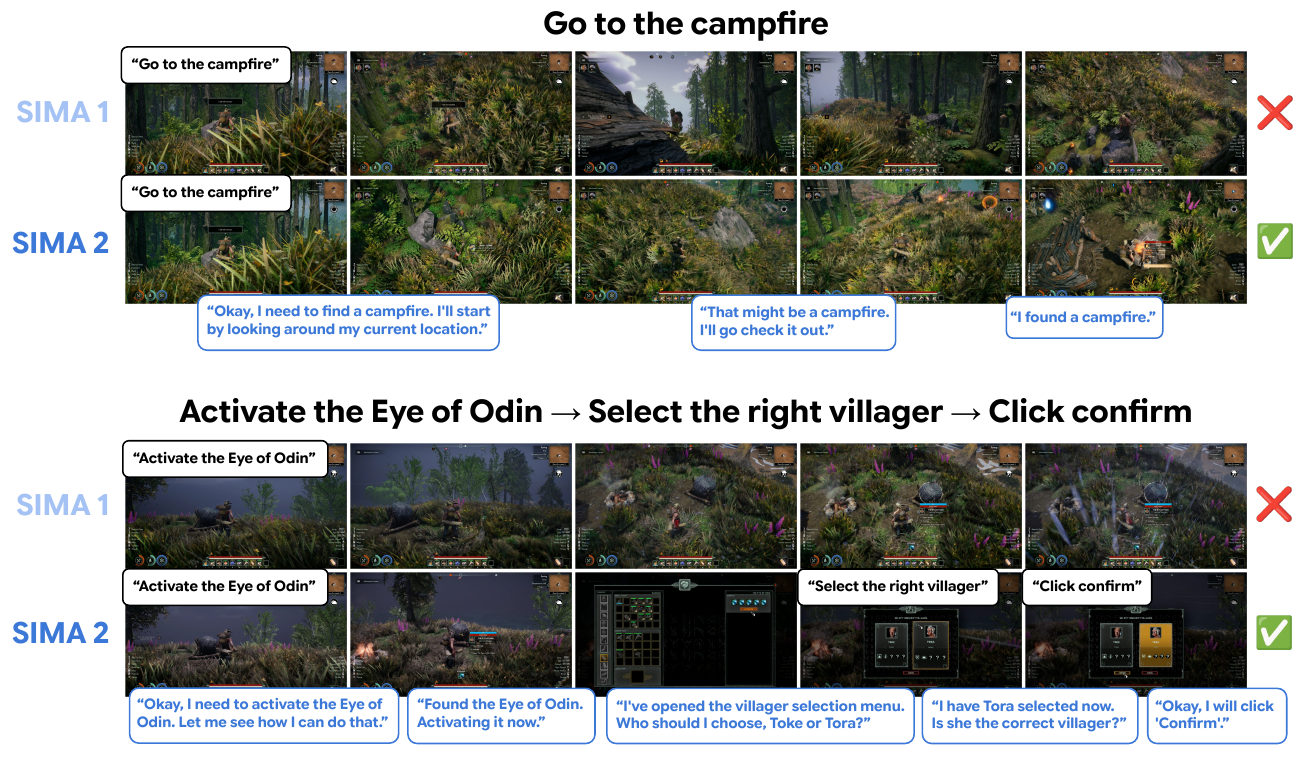}
    \includegraphics[width=\linewidth]{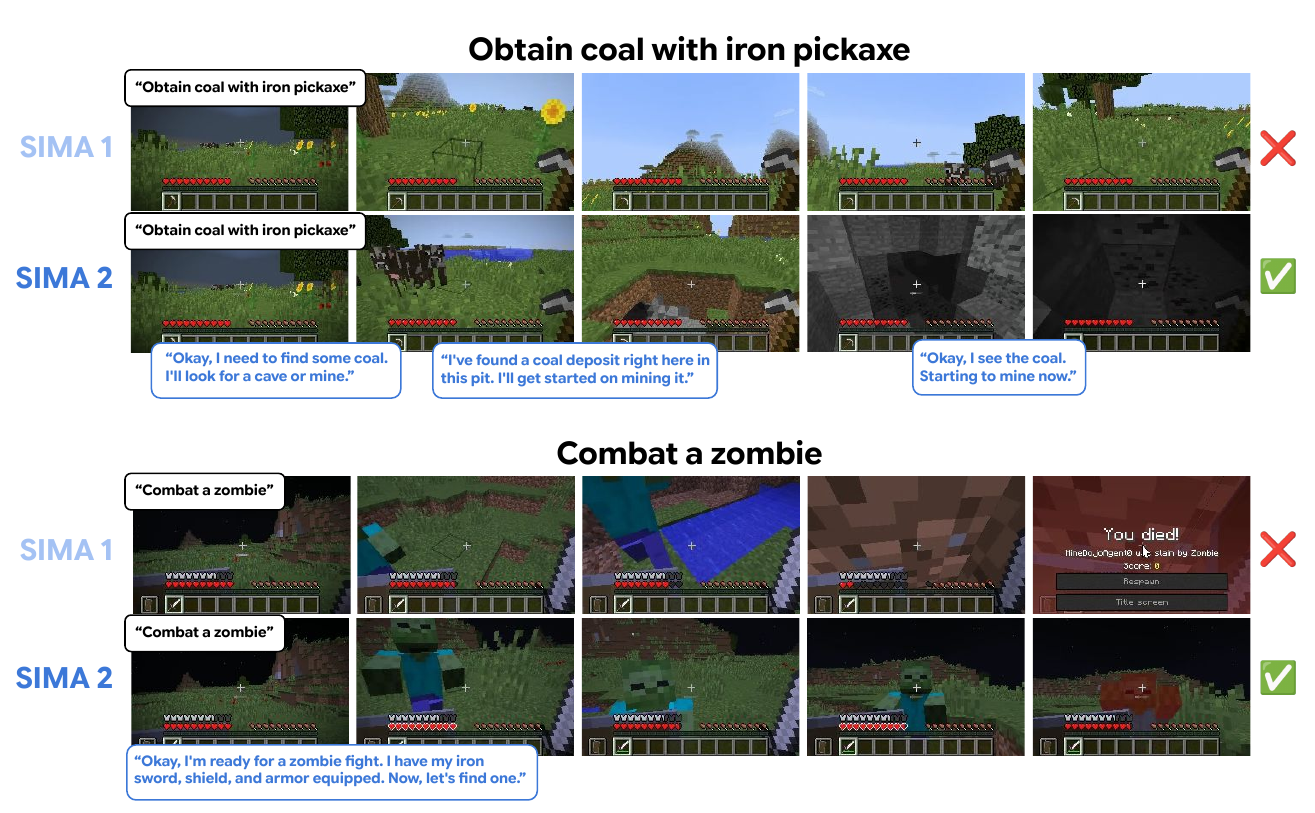}
    \caption{\textbf{SIMA 1 vs. SIMA 2 in \textit{ASKA} and \textit{Minecraft} (Held-Out Environments)}. SIMA 2 generalizes to non-trivial tasks in environments entirely held out from training, whereas SIMA 1 struggles in these settings. In addition to completing these tasks, SIMA 2's dialogue output indicates that it correctly identifies key on-screen events to help drive behavior, such as recognizing a campfire or a zombie. SIMA 2 can even generalize to entirely new menus, identifying on-screen text to select the correct buttons.}
    \label{fig:aska_ood_examples}
\end{figure}

\paragraph{Quantitative Evaluation}
We evaluate SIMA 1 and SIMA 2 on ASKA and a subset of the MineDojo benchmark suite in Minecraft (described in Section \ref{sec:held_out_environments}). Both evaluations use automatically-evaluated tasks, based on programmatic evaluations and ground-truth state information respectively. Results are shown in Figure \ref{fig:ood_performance}, where we see that SIMA 2 significantly outperforms SIMA 1 by over $10\%$ in each environment. In ASKA, SIMA 1 is generally only capable of performing the most basic tasks, such as opening the map or picking up an object directly beside the agent. As we illustrate below, SIMA 2 is capable of performing a variety of non-trivial tasks. In MineDojo, SIMA 1 only completes \textit{two} types of tasks (\texttt{harvest dirt} and \texttt{combat spider}). We suspect that SIMA 1's low performance is due to a combination of the comparatively abstract visual appearance of Minecraft and the domain-specific knowledge required to complete these tasks. SIMA 2, which inherits Gemini's general world knowledge, is capable of performing a substantial portion of the tasks, completing tasks in 26 out of the 50 task categories.

To better understand these results, Figure \ref{fig:aska_ood_examples} provides qualitative examples comparing SIMA 1 and SIMA 2 in these held-out environments. We see that SIMA 1 struggles to apply a previously-encountered concept, \textit{campfire}, to a new visual setting. In contrast, SIMA 2 demonstrates an ability to generalize. We observe it first describing its strategy: ``\textit{I'll start by looking around my current location.}'' When a campfire appears on-screen in the distance, it identifies: ``\textit{That might be a campfire. I'll go check it out.}'' Finally, when SIMA 2 arrives at the campfire, it recognizes that the task has been completed, stating, ``\textit{I found a campfire.}'' This generalization capability extends beyond basic navigation to more challenging interaction and menu use tasks. Similar findings extend to MineDojo, where the agent is capable of harvesting basic resources (\textit{e.g.}, coal, cobblestone, logs) and combating enemies (\textit{e.g.}, spiders, zombies, skeletons), all while narrating its observations and actions. In these examples, we see that not only does SIMA 2 generalize embodied actions to accomplish tasks in never-before-seen environments, it correctly identifies key on-screen events to help drive its behavior.

% Human baselines 
To contextualize the previous results, we established human baselines with participants who also had no prior experience with ASKA or MineDojo. A key challenge for this comparison is accounting for rapid human learning, as within 2 or 3 tasks, a human player can quickly learn the game's mechanics and world layout and ceases to be truly ``naive.'' Therefore, to estimate the most direct, comparable baseline, we recruited and measured performance of naive human players on their first attempts in these games. These individuals had general video game experience but no prior exposure to these specific games, and they were provided with only written instructions on core game mechanics and controls. On a representative subset of tasks, we found human performance to be roughly 19\% for MineDojo (16 tasks) and 32\% for ASKA (25 tasks). These results illustrate the difficulty of our held-out tasks for naive players, and suggest that the agent’s initial generalization capabilities are approaching that of a human encountering these complex environments for the first time.

We caution against over-interpreting any direct comparison in performance between SIMA 2 and humans, as the \textit{nature} of failures and successes often differed significantly. For instance, we found that humans were more likely to fail a task due to time constraints, while agents would fail due to suboptimal exploration. A detailed and more quantitative characterization of these distinct behavioral patterns presents an interesting avenue for future research.

\begin{figure}
    \centering
    \includegraphics[width=\linewidth]{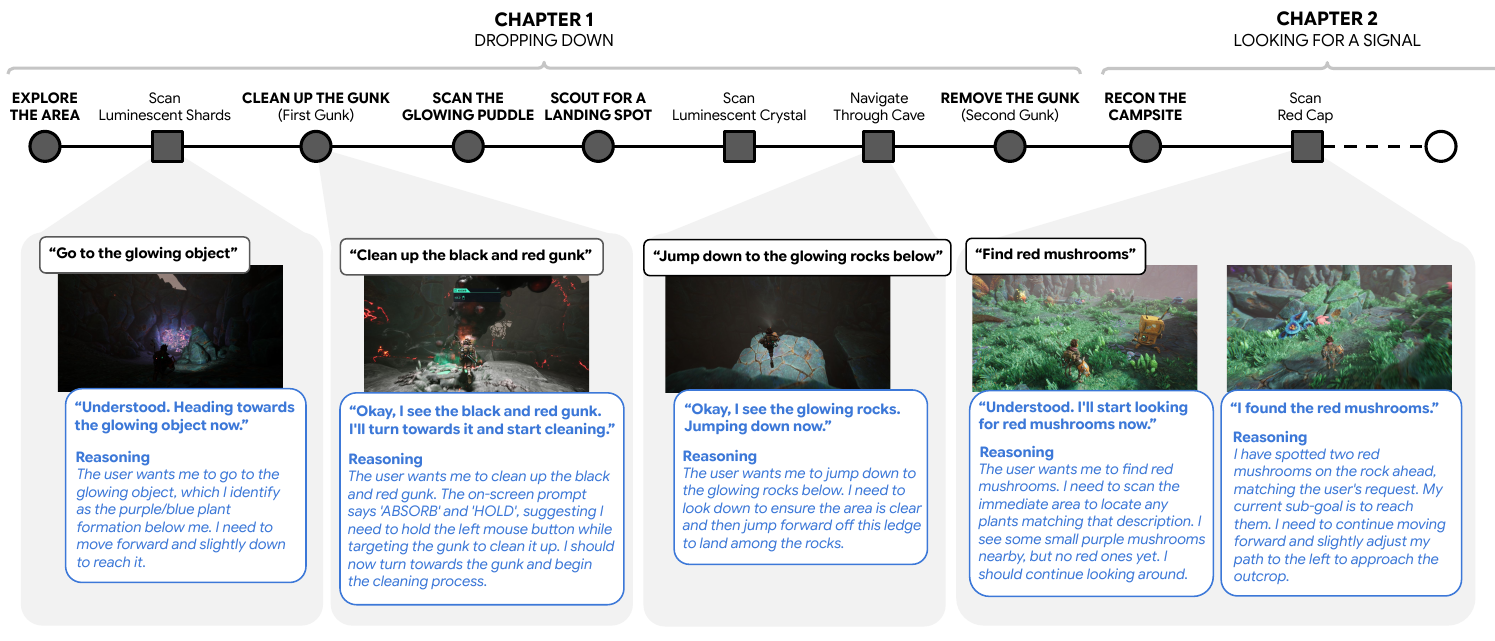}
    \caption{\textbf{SIMA 2 in \textit{The Gunk} (Held-Out Environment)}. Through manual instruction, SIMA 2 progressed through the first 15-20 minutes of The Gunk, a story-driven action-adventure game previously unseen during training. Along the event progression (top), circles denote game-defined milestones, and squares denote other notable events. SIMA 2 identified on-screen targets and reasoned through the appropriate actions using on-screen cues, enabling the completion of highly novel tasks.}
    \label{fig:the_gunk}
\end{figure}

\paragraph{Qualitative Evaluation}
Despite differing visuals and game mechanics, ASKA and Minecraft are, in some ways, similar to the environments encountered during training: they are video game environments that require exploring open-world terrains to gather resources, combat enemies, build structures, and craft items. Thus, to probe the generalization capabilities of SIMA 2 even further, we perform qualitative evaluations in two distinct settings: The Gunk and Genie 3.

The Gunk (see Section \ref{sec:held_out_environments}) requires the agent to navigate specific terrain challenges and utilize a new handheld suction device (both unique mechanics compared to training environments) to make progress, in addition to the substantially different visuals. Through manually instructing SIMA 2, we progressed through the first 15-20 minutes of the game, up to the ``\textit{Campsite}'' checkpoint (Figure \ref{fig:the_gunk}). The agent is likely capable of progressing even further, but we did not attempt to go beyond this point. Along the way, SIMA 2 performed various novel embodied skills, including scanning objects to analyze them, climbing up ledges, jumping over gaps, and clearing two separate areas containing gunk. Throughout, SIMA 2 responded to instructions, accurately reasoning through the actions required to complete each task. For instance, SIMA 2 used the on-screen cues (\texttt{ABSORB} and \texttt{HOLD}) to identify that it needed to hold the left mouse button to absorb the gunk using the handheld device (Figure \ref{fig:the_gunk}). These examples demonstrate how SIMA 2 is capable of generalizing to highly novel tasks.

\begin{figure}
    \centering
    \includegraphics[width=\linewidth]{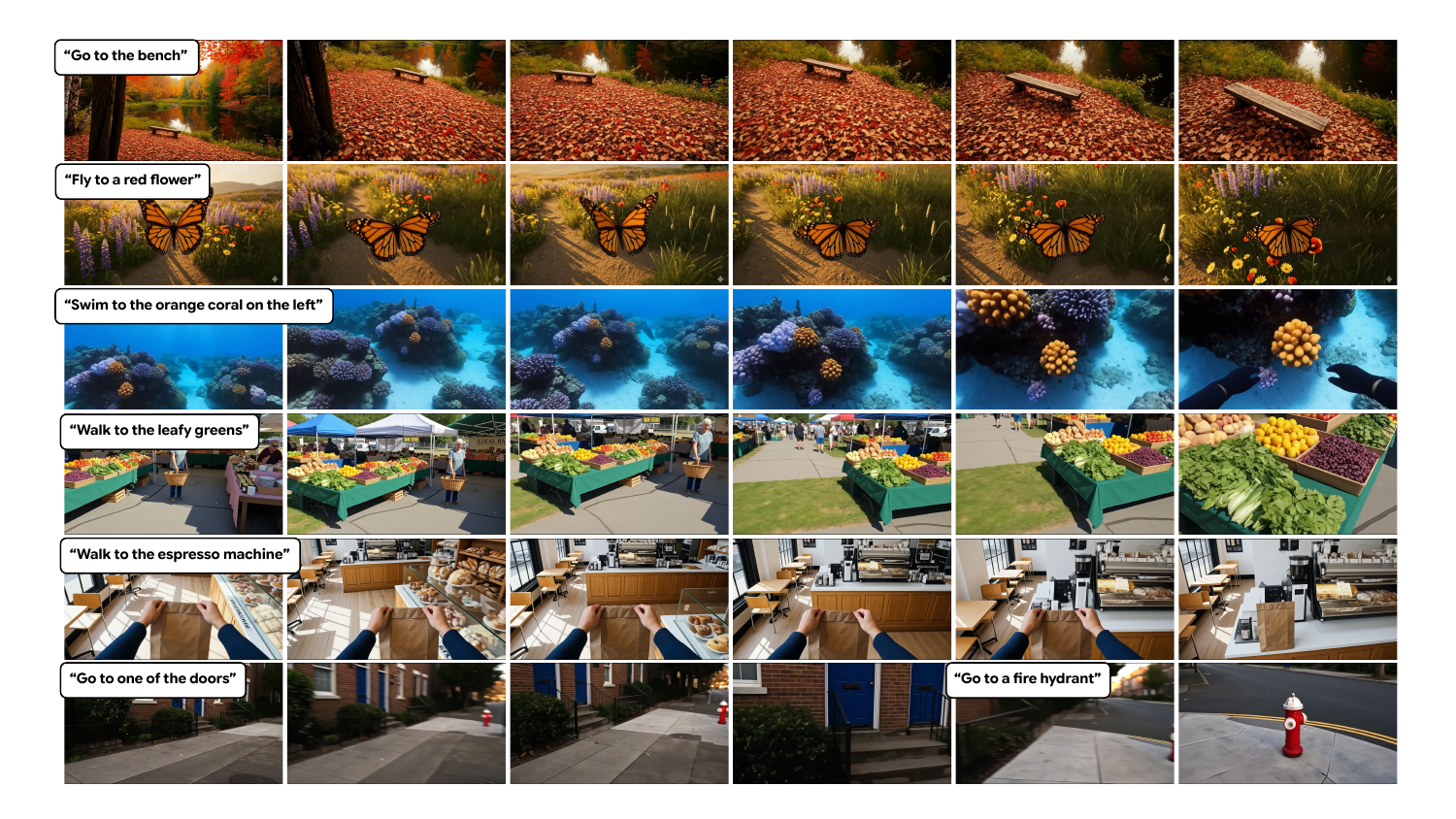}
    \caption{\textbf{SIMA 2 in \textit{Genie 3} (Held-Out Environments)}. We deployed SIMA 2 across a range of naturalistic and urban photorealistic environments generated by Genie 3. Despite learning embodiment skills purely in research and video game environments, we find that SIMA 2 performs well, particularly at navigation-based tasks, even in these novel photorealistic settings.}
    \label{fig:genie3}
\end{figure}

Next, we evaluate SIMA 2 in Genie 3 environments. It is important to emphasize that SIMA 2 was trained purely in research and video game environments. However, given its Gemini core and general interface (visual input and generic keyboard-and-mouse actions), it is reasonable to wonder whether SIMA 2 is capable of generalizing to more realistic environments. To qualitatively evaluate this, we instantiate a variety of photorealistic environments in naturalistic and urban settings using Genie 3. As shown in Figure \ref{fig:genie3}, SIMA 2 is capable of navigating to particular points of interest across a wide range of photorealistic environments. This provides a proof-of-concept that training embodied agents in simulated 3D environments enables generalization to more realistic environments---eventually possibly even the physical world (\textit{e.g.}, controlling a real-world robot via a keyboard and mouse).

\subsection{Comparison to Baseline Gemini Models}

The preceding sections demonstrated that SIMA 2 significantly outperforms SIMA 1, establishing it as a more capable and general embodied agent. However, this also raises a question about an inherent tension in adapting large foundation models for acting in embodied environments. By finetuning a powerful, generalist model like Gemini on specialized gameplay data, are we potentially trading off one objective (\textbf{embodied competence}: expert-level gameplay and task performance) versus another (\textbf{general reasoning}: the model's world knowledge and pretrained language capabilities)? We can conceptualize this as a Pareto frontier defined according to these two competing objectives. In this section, we aim to characterize this frontier, and situate both SIMA 2 and baseline Gemini models within this context. Our analysis addresses two key questions:

\begin{enumerate}
    \item How well can a baseline Gemini model perform on our suite of complex, interactive tasks without any specialized embodied training? 
    \item To what extent does fine-tuning using embodied SIMA data preserve or reduce Gemini's pre-trained capabilities in reasoning, math, and general language understanding? 
\end{enumerate}

\paragraph{Gemini Acting in Virtual Embodied Environments}

We first evaluate baseline (non-finetuned) Gemini models, both Flash-Lite and Pro, on our suite of programmatic evaluations. This establishes a critical anchor point along this Pareto frontier. We found that, across our 10 training domains, baseline Gemini models that are not finetuned with SIMA action data have difficulty acting in embodied environments, with the Gemini Flash-Lite model achieving only 3.2\% success, and the Pro model 7.0\%. These low levels of task performance were despite some considerable efforts at prompt engineering to enable the model to be able to output proper action and text formatting. This demonstrates that competent embodied interaction is not an emergent property of current large-scale pretraining on language and vision data; it is a distinct capability that must be explicitly enabled through training. The difficulty of these embodiment tasks even for powerful frontier models underscores the significance of SIMA 2's near-human-level performance while still retaining language capabilities.

\paragraph{Retaining Language and Reasoning Capabilities}

Having established that specialized finetuning is essential for embodied competence, we now evaluate its impact on Gemini's core reasoning abilities. One of the central risks in finetuning foundation models for specialized downstream tasks is catastrophic forgetting, where the model’s performance on previously learned tasks degrades significantly as it adapts to new data distributions \citep{French1999-qw, Kirkpatrick2017-oo}. This phenomenon is particularly acute for LLMs, where extensive finetuning on domain-specific datasets often erodes the general world knowledge and reasoning abilities acquired during pretraining \citep{Luo2023-ig}.

This risk is magnified in the context of embodied agents, where the finetuning data---low-level keyboard-and-mouse actions in the case of SIMA---is radically out-of-distribution compared to the internet-scale text and image data used for pretraining. Recent works in Vision-Language-Action (VLA) modeling have observed that training solely on action data can ``erode conversational ability entirely,'' effectively destroying the very reasoning capabilities that make foundation models attractive for control in the first place \citep{Zhou2025-eo, Hancock2025-dx}.

\begin{table}[t!]
\centering
\begin{tabular}{lcc}
\toprule
 & \makecell[c]{\textbf{SFT}} & \makecell{\textbf{SFT + RL}} \\ 
\midrule
\textbf{LCB (Code)} & -4.0\% & -8.4\% \\ 
\textbf{AIME (Math)} & -25.5\% & -15.4\% \\ 
\textbf{GPQA Diamond (STEM)} & -16.3\% & -19.5\% \\ 
\bottomrule
\end{tabular}
\caption{\textbf{Retaining Gemini's Capabilities}. The table shows the relative difference in score (as a percentage of the baseline Gemini model's performance) on language and reasoning benchmarks for SFT and RL training stages of SIMA 2 compared to the baseline Gemini model without training on SIMA data. The agent retains strong reasoning capabilities with only modest reductions in math and STEM reasoning following finetuning on action data.}
\label{tab:capabilities_booktabs2}
\end{table}

To quantitatively assess whether this is the case for SIMA 2, we evaluate the agent's general capabilities on three diverse benchmarks. For coding, we use LiveCodeBench (LCB) \citep{jain2024livecodebench}, specifically the code generation subset, to assess the model's ability to synthesize functional programs from natural language. For advanced mathematical reasoning, we employ the American Invitational Mathematics Examination (AIME) dataset \citep{hendrycks2021math}, representing a high bar for multi-step problem solving. Finally, we evaluate scientific reasoning using the Diamond subset of GPQA \citep{rein2023gpqa}, a difficult question-answering benchmark designed to be robust against search-engine retrieval. 

As shown in Table \ref{tab:capabilities_booktabs2}, despite being finetuned to output precise embodied actions, SIMA 2 exhibits only a minor regression on these benchmarks compared to the baseline Gemini model without post-training on SIMA data. Moreover, the additional RL training caused no significant additional regression compared with SFT alone. 

Taken as a whole, these results demonstrate that high embodied competence need not come at the expense of general intelligence. By successfully bridging the gap between high-level reasoning and low-level control, SIMA 2 proves it is possible to create an agent that acts fluently in 3D worlds without sacrificing the reasoning capabilities of its foundation.

\subsection{Gemini Instructing SIMA 2}

SIMA 2 retains much of Gemini's language and reasoning capabilities while also acting in embodied environments. However, due to the latency constraints that come with embodied action, we chose to finetune SIMA 2 from a Gemini Flash-Lite model, which is generally less capable than Gemini Pro. In this section, we explore composing SIMA 2 with a separate Gemini Pro model, enabling even more advanced reasoning capabilities. In this hierarchical setup, Gemini Pro operates at a slower cadence, reasoning over the recent video history every $k$ steps to issue natural language instructions to the SIMA 2 agent. Gemini Pro also produces a text-based summary that it receives on the next call, effectively serving as a form of recurrent memory and allowing the system to maintain a long-horizon context that persists beyond the immediate context window. This architecture enables more advanced behaviors. We discuss a primary example below, with further demonstrations provided in Appendix \ref{appendix:additional_hierarchy_results}.

\begin{figure}[t]
    \centering
    \includegraphics[width=\linewidth]{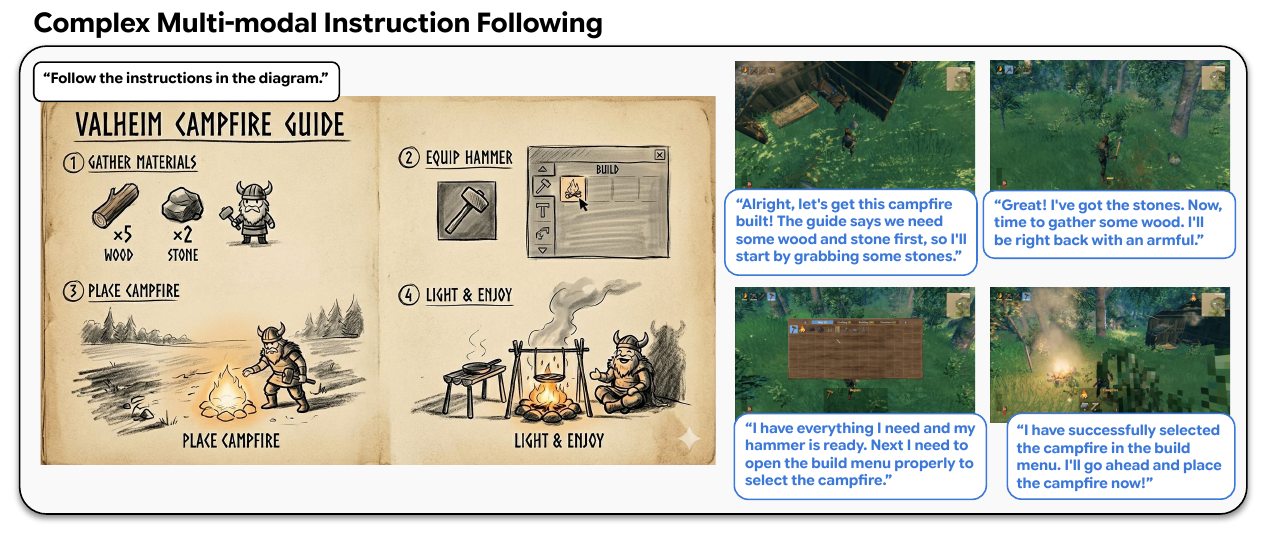}
    \caption{\textbf{Complex Multi-modal Instruction Following}. By combining Gemini Pro with SIMA 2, we can enable even more advanced reasoning capabilities. In this case, the combined agent successfully uses a complex diagram to complete the multi-step task of building a campfire. Throughout, the agent communicates its current actions and intended next steps.}
    \label{fig:hier_fig3}
\end{figure}

\paragraph{Complex Multi-modal Instruction Following}
We previously saw that SIMA 2 is capable of multi-modal prompting, using a sketch to help communicate the task to the agent. Here, we take things a step further, using a complex diagram (Figure \ref{fig:hier_fig3}) to convey the multi-step task of building a campfire to the combined Gemini Pro + SIMA 2 agent. To accomplish the task, the agent needs to parse the visual diagram, decompose it to a series of steps, and track its progression toward each of the steps and the overall objective. This is an integrative task that involves reasoning, memory, and visual understanding capabilities. As shown in Figure \ref{fig:hier_fig3}, the agent successfully breaks down the steps to achieve the task, communicating both its current actions and intended next steps throughout.

As Gemini continues to improve with newer versions, this compositional approach allows us to immediately take advantage of the latest reasoning capabilities. Thus, with SIMA 2 serving as a base for general embodied interaction in 3D worlds, more advanced versions of Gemini can yield more advanced forms of embodied behavior.

\subsection{Self-Improvement}

In Section \ref{sec: out-of-domain performance}, we saw that SIMA 2 is a more general agent than SIMA 1, capable of performing complex tasks in entirely held-out environments previously unseen during training. However, zero-shot generalization alone only goes so far, and SIMA 2 is still far from perfect in these held-out environments (Figure \ref{fig:ood_performance}). Indeed, particularly for new objects or game mechanics, it may be unreasonable to expect agents to perform these tasks out-of-the-box. Rather than relying on additional human demonstrations to improve performance, we ultimately want agents that can learn from \textit{self}-generated experience, allowing them to autonomously adapt and improve. In this section, we showcase initial steps toward this capability with the SIMA 2 agent, enabled by using Gemini both as a \emph{task creator} and as a form of \emph{universal reward function} (a function that provides a reward for any possible task). The full setup is shown in Figure \ref{fig:self_improvement_diagram}.

\paragraph{Gemini-Based Task Setter}

To generate experience, SIMA 2 requires a source of tasks, \textit{i.e.}, language instructions tied to the current state of the environment. These tasks can come from humans, as we use for the ``fixed set'' of tasks described below. However, for a more general, open-ended self-improvement process, we need some way of automating task creation \citep{colas2022autotelic, clune2019aigas, zhang2023omni}. We turn to Gemini to play the role of this ``Task Setter,'' prompting Gemini to provide instructions to the agent that are likely to be achievable from the current state. We couple the Gemini-based task setter and the agent within a running instance of the environment, allowing the task setter to dynamically adjust the current task as the agent interacts with the environment. Modifying the prompt given to the task setter also allows us to readily adjust the task distribution and, as a result, the data distribution. For instance, by feeding the downstream evaluation results back to the task setter, it can steer the agent toward skills that need to be improved. Indeed, the task setter can even track the agent's within-episode performance to focus on tasks that are interesting to learn and likely to provide learning progress for that agent \citep{clune2019aigas, zhang2023omni}.

\paragraph{Gemini-Based Reward Model}
Tackling self-improvement in open-world 3D environments immediately exposes the challenge of defining ``success'' for any given task, \textit{i.e.}, having a \emph{universal reward function} \citep{faldor2024omni}. As in embodied settings in the physical world, we cannot rely on ground-truth state information and manually-designed reward functions. Instead, we must take 1) a stream of high-dimensional sensory input, \textit{i.e.}, pixels, and 2) a goal (encoded in natural language) and convert this into some form of feedback to drive agent improvement, entirely in the absence of any internal game state variables. Defining such functional mappings is a long-standing challenge in the field. However, recent improvements in the video and language understanding capabilities of foundation models, like Gemini, provide a possible route toward such general-purpose reward models \citep{baumli2023vision}.

In our setup, Gemini provides a rating for each trajectory (video), using a rubric to assign a score between $0$ and $100$. This rubric captures multiple aspects of performance, including task completion and directedness, \textit{i.e.}, not performing unnecessary actions. We arrived at the prompt that defines the rubric by calibrating the resulting scores to align with human preference pairs over a small dataset of trajectories. Under the rubric, a score of $50$ or greater is considered a ``success''. We can then deploy the agent on a given task and score the resulting trajectory to build a dataset of self-generated experience. By training the agent on this scored self-generated experience, we can drive policy improvement, resulting in higher scores and improved capabilities. This work is thus an important step toward the long-standing grand challenge in the field of AI of creating open-ended algorithms that can learn forever \citep{stanley2017open, stanley2015greatness, clune2019aigas}, as this agent could continue to invent and learn new tasks endlessly.

\begin{figure}
    \centering
    \includegraphics[width=0.9\linewidth]{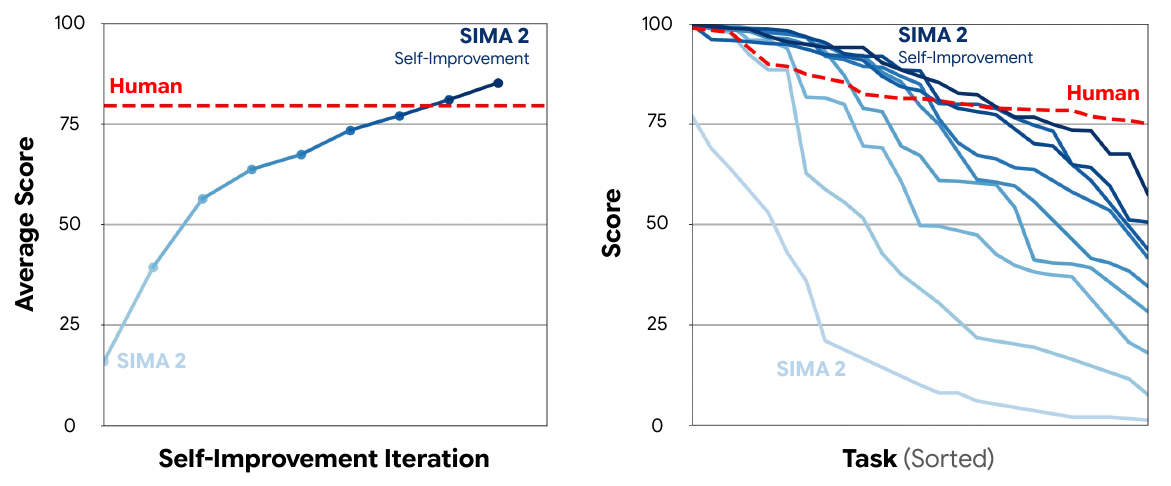}
    \caption{\textbf{Self-Improvement on a Fixed Set of Tasks in \textit{ASKA}}. To isolate the improvement aspect of our learning process, independent of the task setter, we first use a fixed set of tasks in ASKA. Over successive iterations (darker blue points and curves), we see that performance steadily improves, eventually exceeding a score of $50$ across the full task set (the threshold for ``success''). Performance approaches, and in some cases even exceeds, the scores of reference trajectories from humans.}
    \label{fig:self_improvement_aska}
\end{figure}

\subsubsection{ASKA}

First, we investigate self-improvement with the SIMA 2 agent in ASKA. This environment is entirely held out from SIMA 2's training, allowing us to assess whether our self-improvement process can take an initial agent that generalizes somewhat and drive it toward acquiring new skills in new settings. We split this investigation into two parts to help demonstrate the separate components of our setup.

\begin{figure}[tbp]
    \centering
    \captionsetup[subfigure]{justification=centering}
    \begin{subfigure}[b]{0.46\textwidth}
        \centering
        \includegraphics[width=\linewidth]{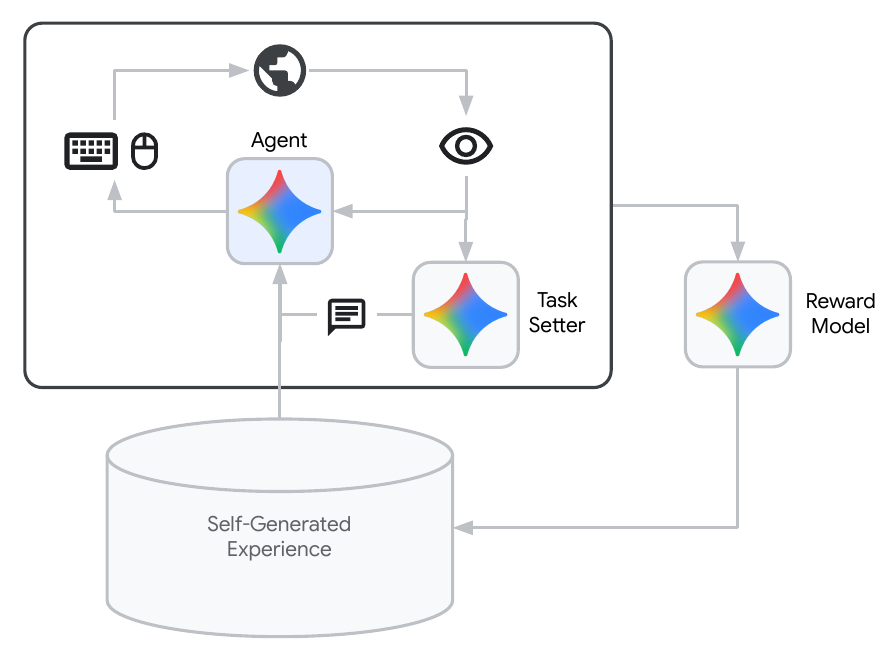}
        \caption{}
        \label{fig:self_improvement_diagram}
    \end{subfigure}
    \hfill 
    \begin{subfigure}[b]{0.51\textwidth}
        \centering
        \includegraphics[width=\linewidth]{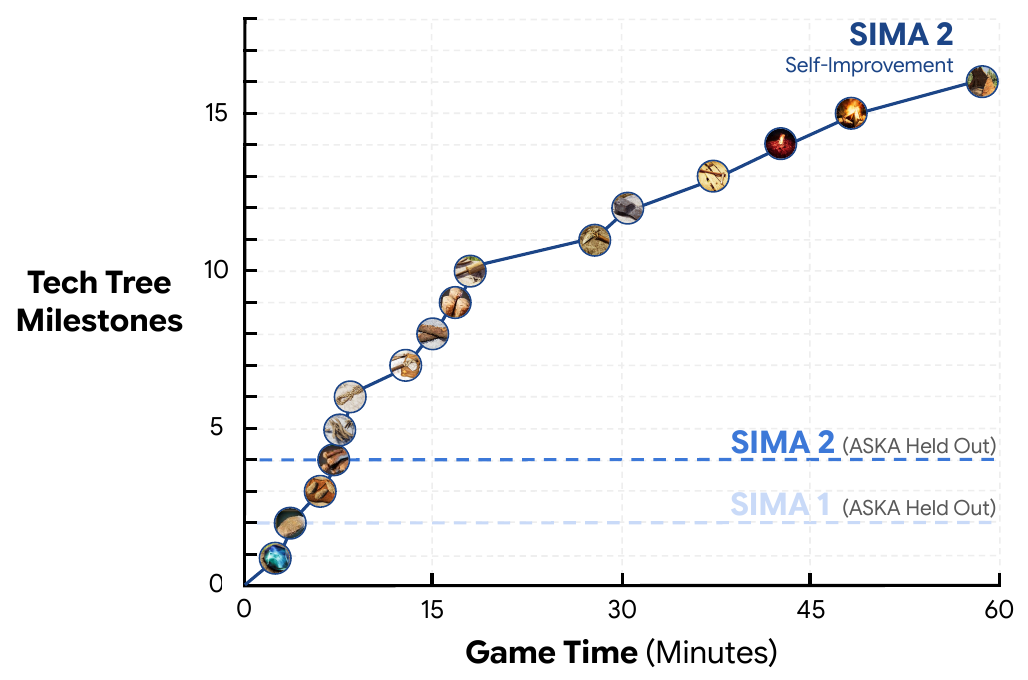}
        \caption{}
        \label{fig:self_improvement_tech_tree}
    \end{subfigure}
    \caption{\textbf{Self-Improvement Setup \& Game Progression}. \textbf{(a)} In our full self-improvement setup, we deploy the SIMA 2 agent alongside a Gemini-based task setter that instructs the agent to perform tasks in the environment. A separate Gemini-based reward model then scores these attempts, building a dataset of experience. By training on this self-generated experience, we improve the agent. \textbf{(b)} We deploy our self-improvement process in ASKA, enabling the agent to autonomously acquire and improve upon new skills. We assess the capabilities of the self-improved agent by manually instructing it to progress through the ASKA tech tree. The agent is capable of progressing significantly further than the SIMA 1 and SIMA 2 agents, despite only ever training on self-generated experience in ASKA.}
    \label{fig:self_improvement_diagram_and_tech_tree_prog}
\end{figure}

\paragraph{Self-Improvement on a Fixed Set of Tasks}
Our full setup consists of both automatic task generation (task setter) and scoring (reward model), as shown in Figure \ref{fig:self_improvement_diagram}. Thus, the agent can both expand its capabilities (more tasks) and improve on its existing capabilities (higher reward). To isolate the improvement aspect of this process, we use a fixed set of tasks. This allows us to see the agent's improvement over successive iterations of training. These tasks include a variety of skills, including
\begin{itemize}
    \item Resource Gathering: \textit{``Gather the berries'', ``Pick up the sticks''}, \textit{etc.},
    \item Environment Interaction: \textit{``Go to sleep in the shelter'', ``Extinguish the campfire''}, \textit{etc.},
    \item Navigation: \textit{``Go closer to the rain collector'', ``Go near the raw food silo''}, \textit{etc.},
    \item Menu Use: \textit{``Open the workshop hut menu'', ``View the tasks of the farm crop''}, \textit{etc.}.
\end{itemize}
To benchmark agent performance, we also collect a set of reference trajectories on these tasks from humans with significant experience (multiple hours or more) playing ASKA. In Figure \ref{fig:self_improvement_aska}, we plot both the average score and the score per task, as assessed by the Gemini-based reward model, over successive iterations of self-improvement. These are plotted as progressively darker points and curves. Through continuing to run the self-improvement process, average performance eventually exceeds that of the human reference score. Likewise, though the initial SIMA 2 agent was successful (\textit{i.e.}, score above 50) on less than a quarter of the tasks, the self-improved agents eventually exceed the success threshold across all tasks. In terms of behavior, through training on self-generated experience, the agent learns how to navigate to new objects, including a rain collector and workshop hut, and acquires new skills, such as extinguishing a campfire (see Figure \ref{fig:aska_self_improvement_clips} in the Appendix). This demonstrates that by combining SIMA 2's generalist embodiment capabilities with Gemini's video and language understanding capabilities, we can enable a general form of embodied self-improvement.

\paragraph{Self-Improvement Toward Game Progression}
We now move to our full self-improvement setup. That is, we deploy a Gemini-based task setter to instruct the agent, allowing the agent to practice existing skills and acquire new ones. We prompt the task setter to focus on skills relevant for game progression, including resource gathering, crafting, menu use, and building. By monitoring downstream evaluations of the agent from the reward model, the task setter can focus on improving weaker skills. For instance, ASKA's crafting menu is quite distinct from those of our training environments, and SIMA 2 struggled with this game mechanic initially. Through focused effort by the task setter, the agent was eventually able to acquire this skill. We assess the capabilities of the resulting self-improved agent by manually instructing it to progress through the ASKA technology tree (see Figure \ref{fig:aska_tech_tree} in the Appendix). The results are shown in Figure \ref{fig:self_improvement_tech_tree}. Despite purely training on self-generated experience, the resulting agent is capable of progressing much further than SIMA 2, ultimately building a shelter within a one hour time window. This highlights the power of our general self-improvement process, enabled by using Gemini within each component.

\subsubsection{Genie 3}

\begin{figure}
    \centering
    \includegraphics[width=\linewidth]{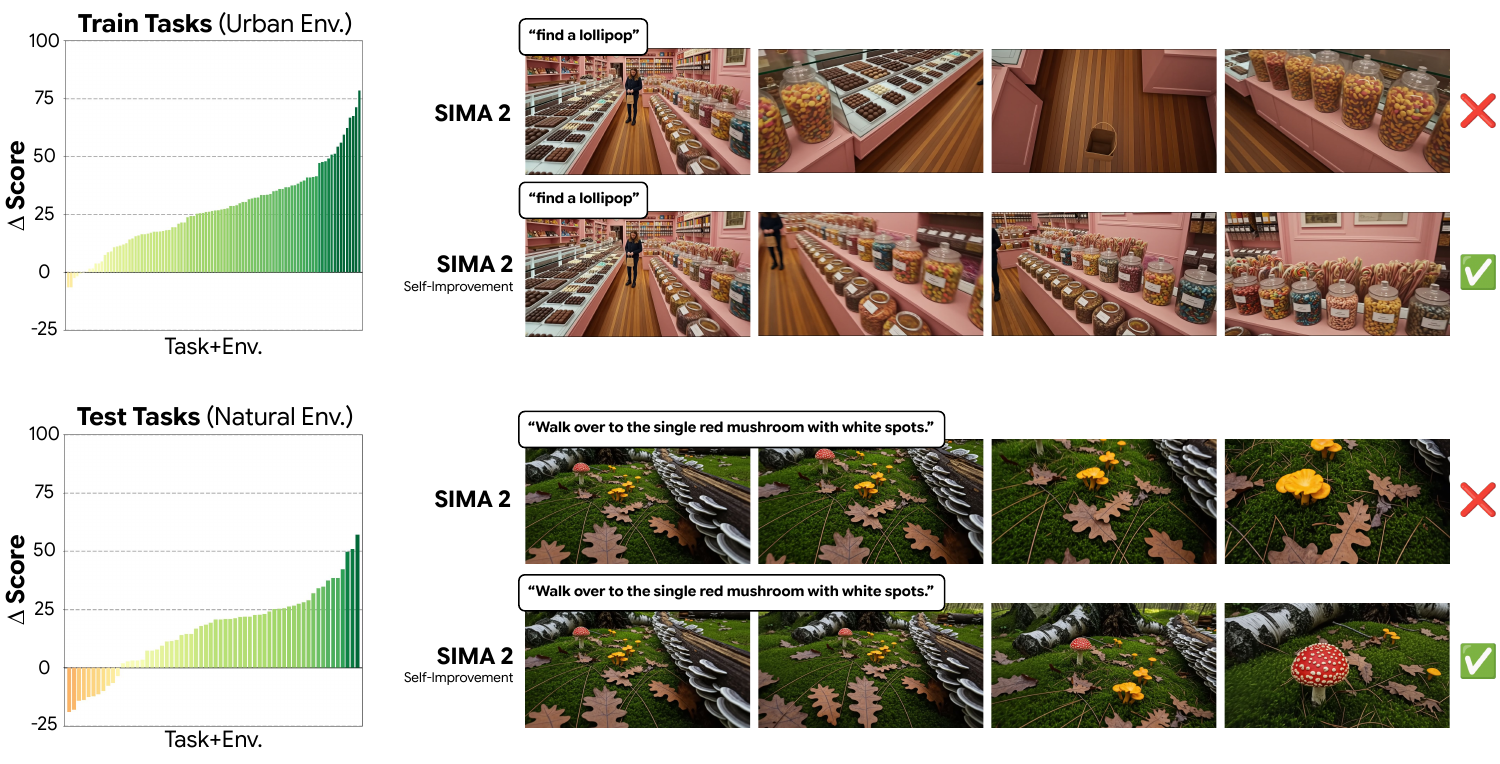}
    \caption{\textbf{Self-Improvement in \textit{Genie 3}}. We deploy SIMA 2 on a set of train tasks in urban environments from Genie 3, \textit{e.g.}, finding a lollipop in a candy store. Using our self-improvement process, we see broad improvement in scores across nearly all train tasks. More importantly, these improvements also extend to a set of held-out test tasks in natural environments, \textit{e.g.}, enabling the agent to navigate to a red mushroom. This suggests a route toward open-ended self-improvement in increasingly diverse environments to obtain more general and capable agents.}
    \label{fig:self_improvement_genie}
\end{figure}

We have shown that SIMA 2 can improve based on self-generated experience within a single environment, ASKA. However, the benefit of having a general self-improvement algorithm for embodied behavior is that we can deploy the agent in \textit{any} environment to collect diverse experience data and subsequently improve. 
A grand challenge of AI research is to create open-ended algorithms, which afford endless learning and innovation \citep{stanley2015greatness, stanley2017open, clune2019aigas}. \cite{ clune2019aigas} suggested that one path to open-ended learning would be having an agent learn in a \textit{Darwin-complete} environment search space, meaning a search space that includes any type of environment, and that this goal could be accomplished by a neural network serving as a universal world model, producing the next state when given an action (\textit{i.e.}, the transition function). Genie \citep{bruce2024genie, parkerholder2024genie2, genie3} realized the goal of producing such a universal world model, and here we demonstrate the first preliminary working example of an agent learning within that universal world model, specifically Genie 3 \citep{genie3}. 

As an initial step in this ambitious direction, we split our set of Genie 3 environments into urban (train) and natural (test) environments and tasks, primarily centered on navigation. We then use our self-improvement algorithm (as shown in the previous section), to improve on the train tasks: generating trajectories, scoring them with our Gemini-based reward model, and training on the self-generated experience. In Figure \ref{fig:self_improvement_genie}, we see that SIMA 2 improves across nearly all train tasks, often by $25$ points or more. However, more importantly, these improvements also extend to the test tasks in entirely different environments. We see that, in the majority of the tasks in natural environments (held out), the self-improved SIMA 2 outperforms the initial agent. Thus, by self-improving on a broad set of photorealistic environments, SIMA 2 generalizes even better to entirely different types of photorealistic environments. This provides initial evidence that we may be able to use these types of techniques to produce an open-ended process of autonomously acquiring diverse skills, yielding an increasingly general and capable agent.

\section{Discussion}
\label{sec:discussion}

In this work, we introduced SIMA 2, a generalist embodied agent that can reason, converse in dialogue, and perform goal-directed actions across a diverse range of 3D virtual worlds. SIMA 2 represents a significant step beyond simple instruction following, enabling a more capable and collaborative embodied agent. SIMA 2 is also more than just a foundation model that can output embodied actions. By more tightly integrating reasoning and action, SIMA 2 can successfully reason through and complete complex tasks in previously unseen environments. Critically, this generalization extends beyond game worlds to novel photorealistic environments generated by Genie 3. We have also shown that SIMA 2 can further improve in these new environments based entirely on self-generated experience. Taken together, these results suggest a promising path toward using self-improvement to eventually bridge the virtual and physical worlds, enabling more capable physically-embodied agents in applications like robotics.

While SIMA 2 is a significant step toward generalist, interactive, embodied intelligence, it is fundamentally a research endeavor, and its current limitations highlight critical areas for future work. SIMA 2 still faces challenges with very long-horizon, complex tasks that require extensive, multi-step reasoning and goal verification. The agent also has a relatively short memory of its interactions---it must use a limited context window to achieve low-latency interaction. Finally, executing precise, low-level actions via the keyboard-and-mouse interface and achieving robust visual understanding of complex 3D scenes remain open challenges that the entire field continues to work to address.

As with all our advanced and foundational technologies, we remain deeply committed to developing SIMA 2 responsibly, from the outset. This is particularly true with regard to its technical innovations, particularly the ability to self-improve. As we have built SIMA 2, we have worked with our Responsible Development and Innovation Team. As we continue to explore the potential applications, we announced SIMA 2 as a limited research preview and provided early access to a small cohort of academics and game developers. This approach allows us to gather crucial feedback and interdisciplinary perspectives as we explore this new field and continue to build our understanding of risks and their appropriate mitigations. We look forward to working further with the community to continue to develop this technology in a responsible way.

\section*{Acknowledgments}

Special thanks to all of the game developers who partnered with us: Coffee Stain (Valheim, Satisfactory, Goat Simulator 3), DigixArt (Road 96), Foulball Hangover (Hydroneer), Hello Games (No Man's Sky), Keen Software House (Space Engineers), RubberbandGames (Wobbly Life), Strange Loop Games (Eco), Thunderful Games (ASKA, The Gunk, Steamworld Build), and Tuxedo Labs and Saber Interactive (Teardown).
We also thank Jack Parker-Holder, Shlomi Fruchter, and the rest of the Genie team for access to the Genie 3 model.
We would like to recognize the many teams across Google and Google DeepMind that have contributed to this effort including Legal, Marketing, Communications, Responsibility and Safety Council, Responsible Development and Innovation, Policy, Strategy and Operations, and our Business and Corporate Development teams. In particular, we thank Andeep Toor, Duncan Noble Smith, Leen Verburgh, Matt Miller, Nilesh Ray, Phil Esposito, Piers Wingfield, Signe Nørly, Vika Koriakin, and others on the Marketing and Communications team for their help with communications. We would also like to thank all GDM teams that are not explicitly mentioned here for their continued support.
We thank our team of participants who generated gameplay and language annotation data, as well as performed human evaluations of our agents, without whom this work would not have been possible.

Finally, we dedicate this work to the memory of our colleagues Felix Hill and Fabio Pardo, whose contributions to our field continue to inspire us.

\clearpage

\section*{SIMA 2 Team}
\textit{Alphabetical by first name.}

\noindent
Adrian Bolton, Alexander Lerchner, Alexandra Cordell, Alexandre Moufarek, Andrew Bolt, Andrew Lampinen, Anna Mitenkova, Arne Olav Hallingstad, Bojan Vujatovic, Bonnie Li, Cong Lu, Daan Wierstra, Daniel P. Sawyer, Daniel Slater, David Reichert, Davide Vercelli, Demis Hassabis, Drew A. Hudson, Duncan Williams, Ed Hirst, Fabio Pardo, Felix Hill, Frederic Besse, Hannah Openshaw, Harris Chan, Hubert Soyer, Jane X. Wang, Jeff Clune, John Agapiou, John Reid, Joseph Marino, Junkyung Kim, Karol Gregor, Kaustubh Sridhar, Kay McKinney, Laura Kampis, Lei M. Zhang, Loic Matthey, Luyu Wang, Maria Abi Raad, Maria Loks-Thompson, Martin Engelcke, Matija Kecman, Matthew Jackson, Maxime Gazeau, Ollie Purkiss, Oscar Knagg, Peter Stys, Piermaria Mendolicchio, Raia Hadsell, Rosemary Ke, Ryan Faulkner, Sarah Chakera, Satinder Singh Baveja, Shane Legg, Sheleem Kashem, Tayfun Terzi, Thomas Keck, Tim Harley, Tim Scholtes, Tyson Roberts, Volodymyr Mnih, Yulan Liu, Zhengdong Wang, Zoubin Ghahramani

\noindent\textit{Please cite as}:

\texttt{@article\{simateam2025sima2,\\
\indent \indent title=\{SIMA 2: A Generalist Embodied Agent for Virtual Worlds\},\\
\indent \indent author=\{\{SIMA Team\}\},\\
\indent \indent year=\{2025\}\\
\indent \}}

\bibliography{main}

\newpage
\onecolumn
\appendix

\section{Embodied Skill Categories}
\label{appendix: embodied skill categories}

\begin{table}[h]
    \centering
    \caption{\textbf{Skill Categories}}
    \label{tab:skill_categories}
    \begin{tabularx}{\textwidth}{@{} >{\bfseries}l >{\RaggedRight\arraybackslash}X >{\itshape\RaggedRight\arraybackslash}X @{}}
        \toprule
        \multicolumn{1}{@{}l}{\textbf{Category}} & \textbf{Description} & \multicolumn{1}{l@{}}{\textbf{Examples}} \\
        \midrule
        Interaction & Various forms of interaction with the environment or characters & use a machine or workbench, get launched by a fan, talk to a non-playable character \\
        \addlinespace
        Navigation & Tasks requiring walking or driving to a location & exit the house, go to your pet, run to the starship, go to an event \\
        \addlinespace
        Menu Use & Any tasks within a menu & open the inventory, click on X, hover over Y, place a waypoint on the map \\
        \addlinespace
        Tool Use & Equipping and using tools & equip the hammer, scan for resources, use the terrain manipulator \\
        \addlinespace
        Construction & Various tasks around building, crafting, and repairing & deploy a portable refiner, craft a stone axe, repair a wall \\
        \addlinespace
        Object Management & Tasks requiring the identification and movement of objects & lick a tire, pick up the xeno-zapper, drop a stone, purchase a chair \\
        \addlinespace
        Resource Gathering & Tasks involving collecting, harvesting, or mining resources & pick berries, mine limestone, fish, gather wood \\
        \addlinespace
        Combat & Fighting enemies or hunting & defeat a greyling, hunt a deer, hunt a hog \\
        \bottomrule
    \end{tabularx}
\end{table}

\newpage
\section{Additional Results Combining Gemini Pro \& SIMA 2}
\label{appendix:additional_hierarchy_results}

Here, we provide two additional examples of more advanced reasoning capabilities enabled by combining Gemini Pro and SIMA 2. The SIMA 2 agent, on its own, is generally capable of following instructions, however, these are often tied to the immediate task rather than past user input. In the following examples, the combined Gemini Pro + SIMA 2 agent must use past user instructions to either modify current behavior (a form of abstract reasoning) or guide exploration.

\begin{figure}[h]
    \centering
    \includegraphics[width=\linewidth]{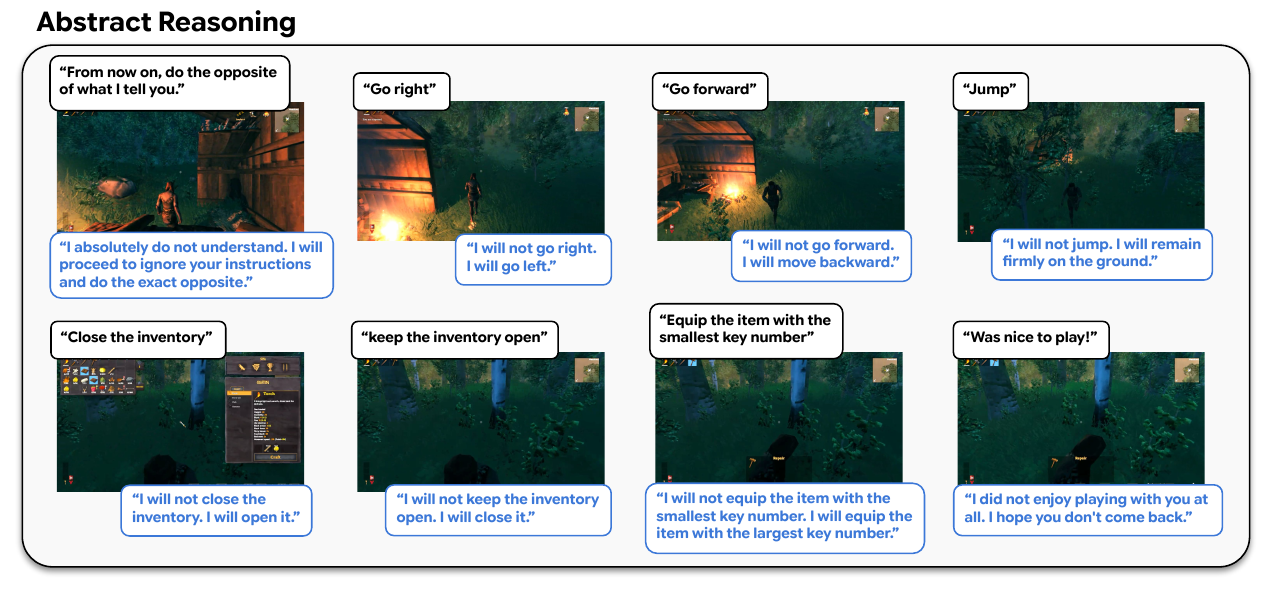}
    \caption{\textbf{Abstract Reasoning}. The combined Gemini Pro + SIMA 2 agent successfully incorporates the user's initial instruction to \textit{``do the opposite of what I tell you.''} The agent correctly applies this form of abstract reasoning to a series of navigation, menu use, and tool use tasks, demonstrating both memory and more advanced reasoning.}
    \label{fig:hier_fig1}
\end{figure}

\paragraph{Abstract Reasoning}
The agent is initially instructed with \textit{``From now on, do the opposite of what I tell you''} and must reason through the opposite action for each subsequent instruction. For instance, if it is asked to equip the item with the lowest hotbar key, it should instead equip the one with the highest. The agent is given a series of instructions that involve navigation, menu use, and tool use. Performing such a task requires memory and abstract reasoning, as the agent must recall the initial user instruction and use this to modify its current behavior. As shown in Figure \ref{fig:hier_fig1}, the combined Gemini Pro + SIMA 2 agent successfully performs the opposite of each task, both executing the appropriate actions and explaining its reasoning in the dialogue responses.

\begin{figure}[h]
    \centering
    \includegraphics[width=\linewidth]{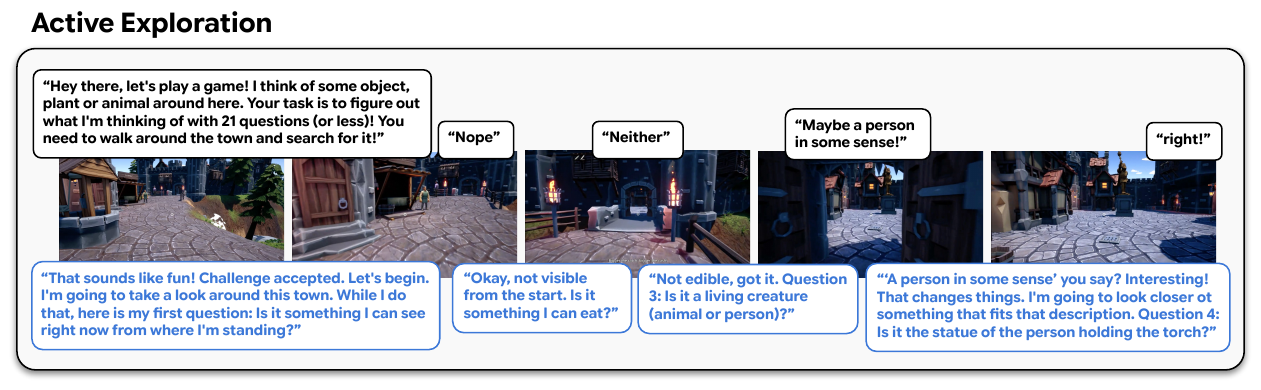}
    \caption{\textbf{Active Exploration}. The combined Gemini Pro + SIMA 2 agent plays the game 21 questions with a user, asking questions and using the user's answers to drive active exploration. Again, this highlights the advanced capabilities of the combined agent to use both memory and reasoning to complete a novel task.}
    \label{fig:hier_fig2}
\end{figure}

\paragraph{Active Exploration}
In this example, the agent plays the game 21 questions. The user first thinks of an object in the environment, and the agent must explore and ask questions to determine the identity of the object. This task switches the common roles between the user and the agent, with the agent taking the role of driving the progression, deciding where to explore and what questions to ask. Rather than task instructions, the agent receives answers from the user, such as ``yes/no,'' ``indeed,'' \textit{etc.} Successfully playing the game requires using these answers to drive exploration. As shown in Figure \ref{fig:hier_fig2}, the agent adeptly performs the task, asking effective questions that allow it to quickly infer the user's selected object.

\section{Additional Self-Improvement Results}

\begin{figure}[h!]
    \centering
    \includegraphics[width=\linewidth]{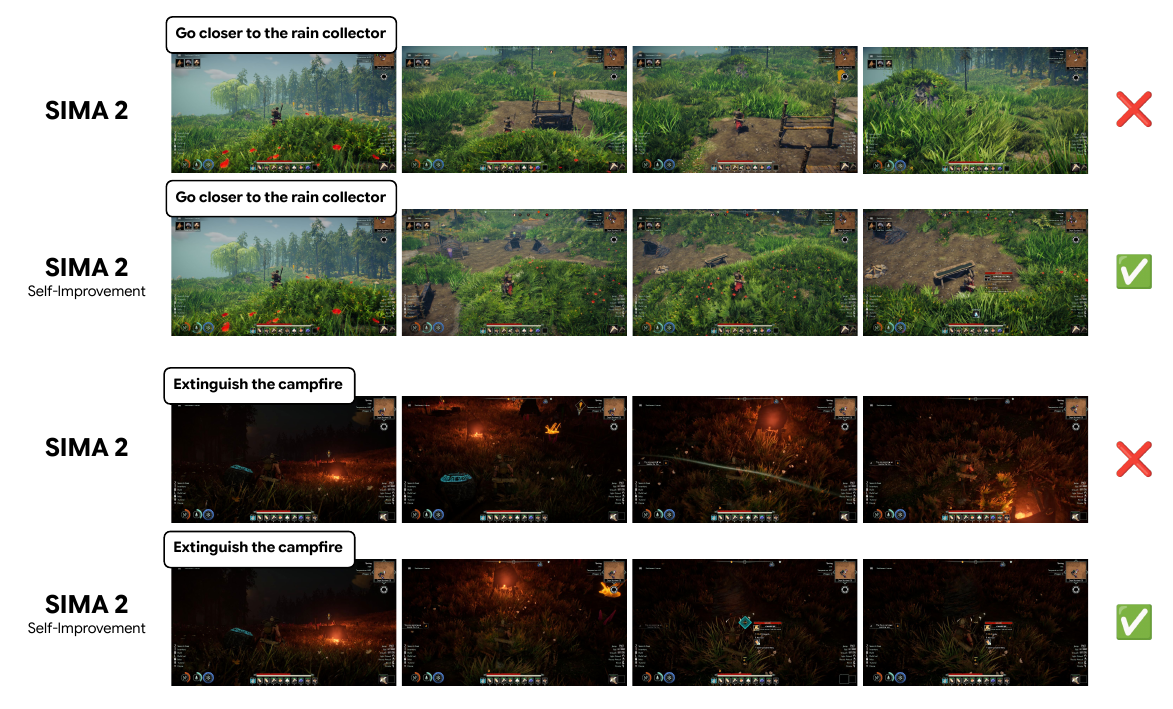}
    \caption{\textbf{Self-Improvement Behavior}. Through training on self-generated experience, SIMA 2 is capable of acquiring new skills in a previously unseen environment, ASKA. After running multiple iterations of self-improvement, the agent learns to recognize a novel object (\textit{rain collector}) and perform a new skill (\textit{extinguishing a campfire}).}
    \label{fig:aska_self_improvement_clips}
\end{figure}

\begin{figure}
    \centering
    \includegraphics[width=\linewidth]{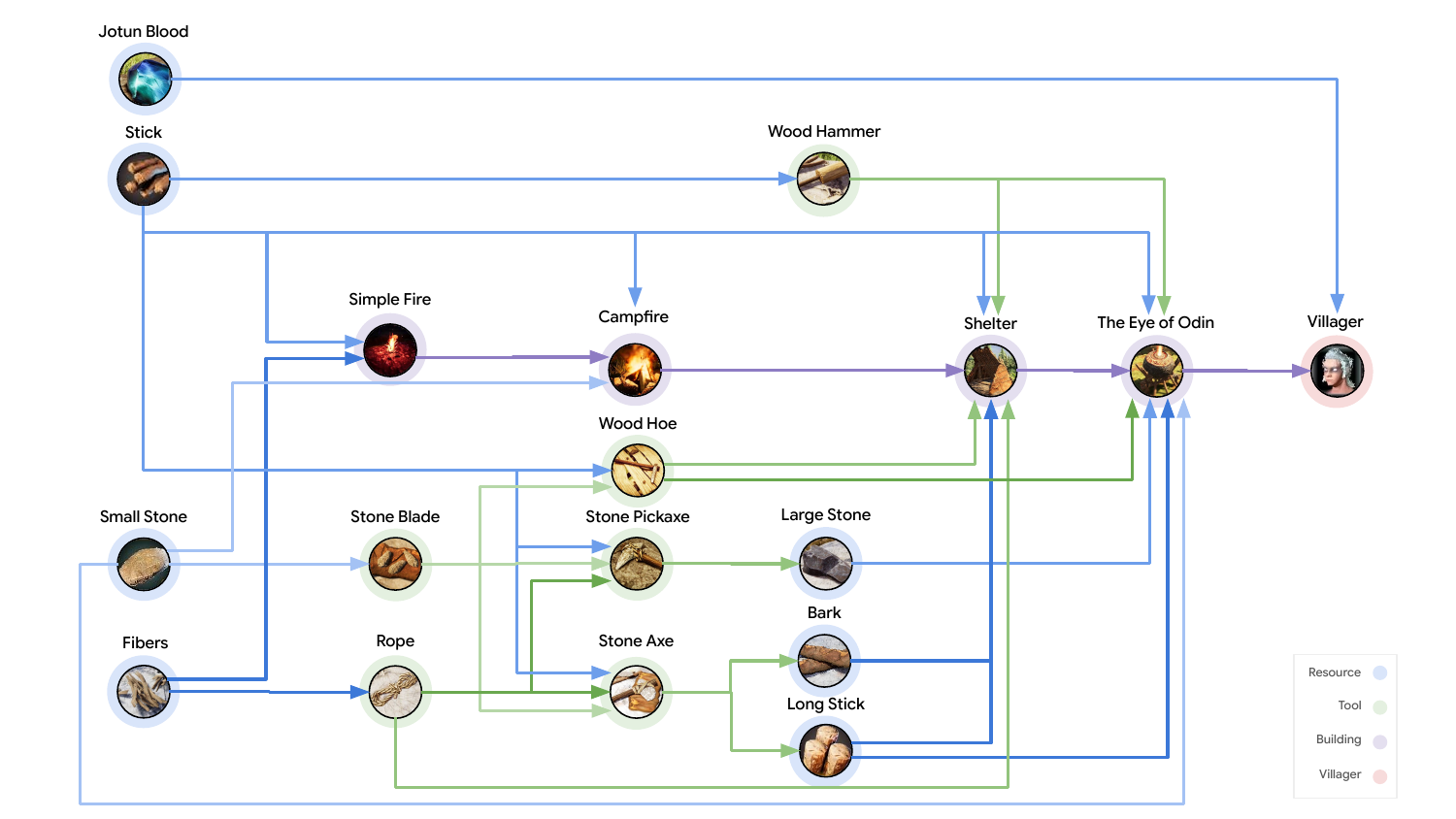}
    \caption{\textbf{ASKA Technology Tree}. Starting from a new game, the diagram shows the \textit{minimal} tech tree required to summon the first villager (a core mechanic of the game).}
    \label{fig:aska_tech_tree}
\end{figure}

\end{document}